\documentclass[10pt,twocolumn,letterpaper]{article}

\usepackage{cvpr}
\pdfoutput=1 
\usepackage{times}
\usepackage{epsfig}
\usepackage{graphicx}
\usepackage{amsmath}
\usepackage{amssymb}
\usepackage{amsmath}
\usepackage{enumitem}
\usepackage[skip=0pt]{subcaption}
\usepackage[normalem]{ulem}
\usepackage{balance} % try to get references balanced on last page
\usepackage{placeins} % try to not embed table in references section
\usepackage{tabularx}
\usepackage[para]{footmisc}
\usepackage[pagebackref=true,breaklinks=true,letterpaper=true,colorlinks,bookmarks=false]{hyperref}

\usepackage{titling}
\cvprfinalcopy % *** Uncomment this line for the final submission

\def\cvprPaperID{8850} % *** Enter the CVPR Paper ID here

\ifcvprfinal\fi
\date{\vspace{-0ex}}
\def\@seccntformat#1{Appendix\ \csname the#1\endcsname\quad}
\makeatother

\newcommand{\rr}{\mathbf{r}}
\newcommand{\btt}{\mathbf{t}}		% also some problem

\newcommand{\vv}{\mathbf{v}}

\newcommand{\RR}{\mathbf{R}}

\newcommand{\SECREF}[1]{{Section~\ref{#1}}}   % spelling of "section"
\newcommand{\FIGREF}[1]{{Figure~\ref{#1}}}   % spelling of "section"%\newcommand{\SECREF}{Section}  % or Sec.

\usepackage{xcolor}
\definecolor{dark_red}{rgb}{0.9, 0, 0}
\definecolor{light_grey}{rgb}{0.8, 0.8, 0.8}
\definecolor{dark_blue}{rgb}{0, 0, 0.7}
\definecolor{dark_green}{rgb}{0, 0.5, 0}

\begin{document}
\setlength{\abovedisplayskip}{0pt}
\setlength{\belowdisplayskip}{0pt}
%%%%%%%%% TITLE
\title{\vspace{-12mm}LipSync3D: Data-Efficient Learning of Personalized 3D Talking Faces from Video using Pose and Lighting Normalization }

\makeatletter
\renewcommand*{\@fnsymbol}[1]{\ensuremath{\ifcase#1\or *\or \dagger\or \ddagger\or
    \mathsection\or \mathparagraph\or \|\or **\or \dagger\dagger
    \or \ddagger\ddagger \else\@ctrerr\fi}}
\makeatother
\newcommand*\samethanks[1][\value{footnote}]{\footnotemark[#1]}
\newcommand*\inst[1]{$^{#1}$}

\author{
Avisek~Lahiri\inst{1,2}\thanks{Equal contribution}\hspace{.05in}\thanks{Work done while an intern at Google}
\and
Vivek~Kwatra\inst{1}\samethanks[1]
\and
Christian~Frueh\inst{1}\samethanks[1]
\and
John~Lewis\inst{1}
\and
Chris~Bregler\inst{1}\\
\and
$^1$Google Research~~~$^2$Indian Institute of Technology Kharagpur\\
{\tt\small \{avisek,kwatra,frueh,jplewis,bregler\}@google.com}
}

\maketitle
\ifcvprfinal\thispagestyle{empty}\fi
\vspace{-2mm}
\begin{abstract}
\vspace{-2mm}
%%%%%%%% ABSTRACT
In this paper, we present a video-based learning framework for animating personalized 3D talking faces from audio.
We introduce two training-time data normalizations that significantly improve data sample efficiency. First, we isolate and represent faces in a normalized space that decouples 3D geometry, head pose, and texture. This decomposes the prediction problem into regressions over the 3D face shape and the corresponding 2D texture atlas. Second, we leverage facial symmetry and approximate albedo constancy of skin to isolate and remove spatio-temporal lighting variations. Together, these normalizations allow simple networks to generate high fidelity lip-sync videos under novel ambient illumination while training with just a single speaker-specific video. Further, to stabilize temporal dynamics, we introduce an auto-regressive approach that conditions the model on its previous visual state. Human ratings and objective metrics demonstrate that our method outperforms contemporary state-of-the-art audio-driven video reenactment benchmarks in terms of realism, lip-sync and visual quality scores. We illustrate several applications enabled by our framework.

\end{abstract}
\vspace{-3mm}
\section{Introduction}
\begin{figure}[t]
\centering
\begin{subfigure}{\columnwidth}
  \centering
  \includegraphics[width=0.9\columnwidth]{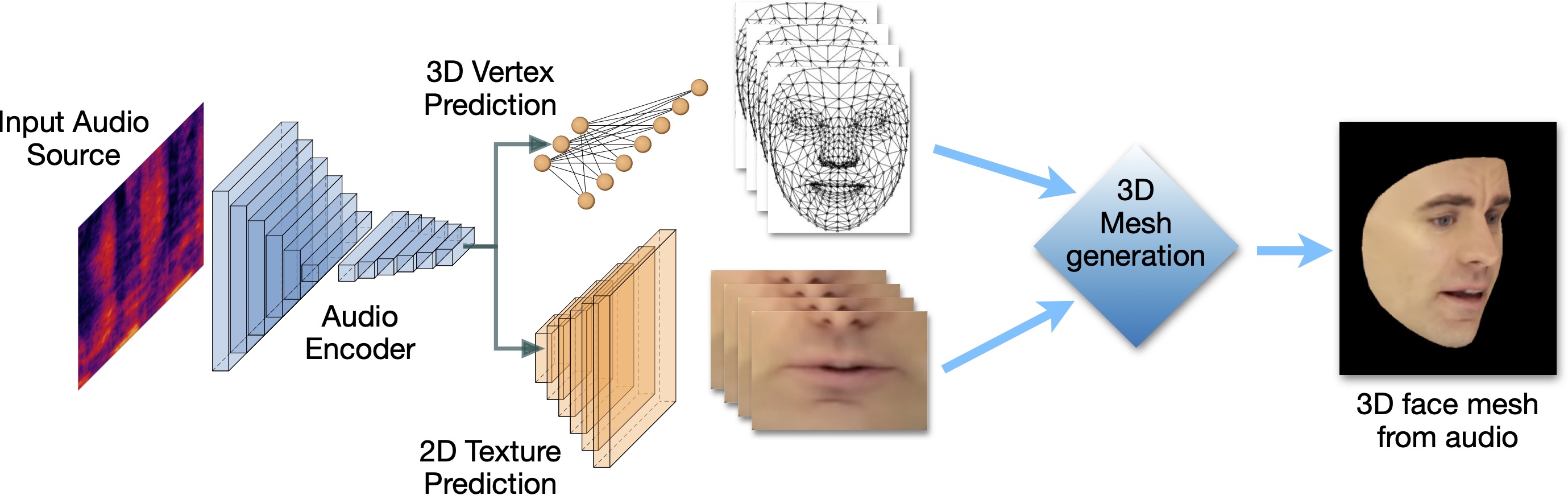} % ARXIV version
  
   \caption{}
   \label{fig_flow_mesh} 
\end{subfigure}
  \par\medskip
\begin{subfigure}{\columnwidth}
  \centering
   \includegraphics[width=0.9\columnwidth]{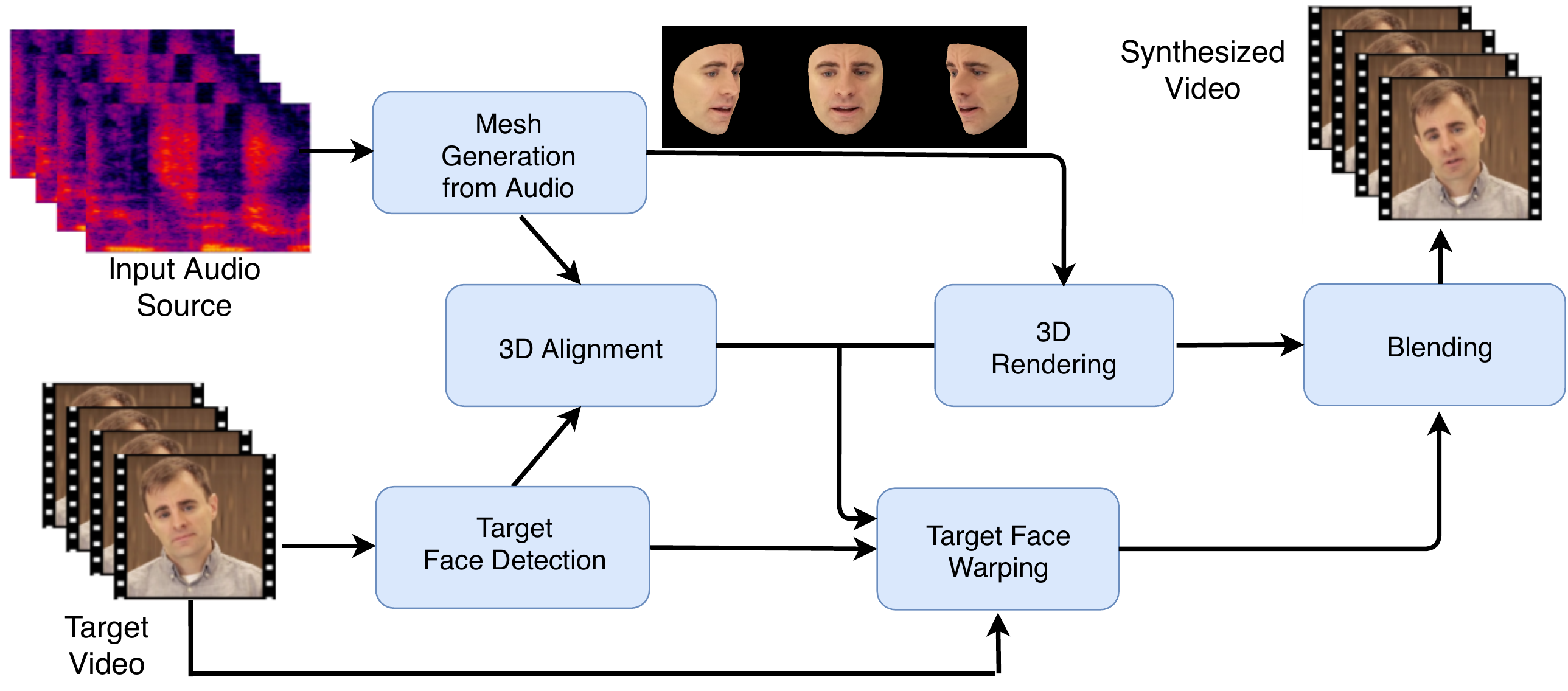}
   \caption{}
   \label{fig_flow_video}
\end{subfigure}
\caption{\scriptsize Flow diagram of our approach to  
(a) generate a dynamically textured 3D face mesh from audio, and (b) insert the generated face mesh into a target video to create a synthesized talking head video from new audio input.}

\label{fig_flow}
\vspace{-1.5em}
\end{figure}

%=====================================================
%WHAT IS THE PROBLEM, WHY IMPORTANT:
``Talking head'' videos, consisting of closeups of a talking person, are widely used in newscasting, video blogs, online courses, etc. 
Other applications that feature talking faces prominently are face-to-face live chat, 3D avatars and animated characters in games and movies. 
% SHORTEN Other modalities, with similar frame composition focusing on faces, include face-to-face live chat and 3D game avatars. 
We present a deep learning approach to synthesize 3D talking faces (both photorealistic and animated) driven by an audio speech signal. We use speaker-specific videos to train our model in a data-efficient manner by employing 3D facial tracking.
The resulting system has multiple applications, including video editing, 
lip-sync for dubbing of videos in a new language, personalized 3D talking avatars in gaming, VR and CGI, as well as compression in multimedia communication.

%WHAT HAS BEEN TRIED BEFORE:
The importance of talking head synthesis has led to a variety of methods in the research literature. Many recent techniques \cite{chen2018lip,chen2019hierarchical,x2face,aaai,wacv,wav2lip} use the approach of regressing facial motion from audio, employing it to deform one or more reference images of the subject. These approaches can inherit the realism of the reference photos, however, the results
do not accurately reproduce 3D facial articulation and appearance
under general viewpoint and lighting variations. Another body of research predicts 3D facial meshes from audio~\cite{wang2011text,edwards2016jali,Karras,cudeiro2019capture}. These approaches are directly suitable for VR and gaming applications. However, visual realism is often restricted by the quality of texturing. Some recent approaches~\cite{obama,thies2020neural,fried2019text} attempt to  bridge the gap by combining 3D prediction with high-quality rendering, but are only able to edit fixed target videos that they train on.

% YOUR SOLUTION
Our work encompasses several of the scenarios mentioned above. We can use 3D information to edit 2D video, including novel videos of the same speaker not seen during training. We can also drive a 3D mesh from audio or text-to-speech (TTS), and synthesize animated characters by predicting face blendshapes. 

Next, we highlight some of our key design choices. 

\textbf{Personalized models:} We train personalized speaker-specific models, instead of building a single universal model to be applied across different people. While universal models like Wav2Lip~\cite{wav2lip} are easier to reuse for novel speakers, they need large datasets for training and do not adequately capture person-specific idiosyncrasies~\cite{bregler1997video}. Personalized models like ours and NVP~\cite{thies2020neural} produce results with higher visual fidelity, more suitable for editing long speaker-specific videos. Additionally, our model can be trained entirely using a single video of the speaker.

%\vspace{0.1in}
%\noindent
\textbf{3D pose normalization:} We use a 3D face detector~\cite{xeno} to obtain the pose and 3D landmarks of the speaker's face in the video. This information allows us to decompose the face into a normalized 3D mesh and texture atlas, thus decoupling head pose from speech-induced face deformations,~\eg lip motion and teeth/tongue appearance.

%\vspace{0.1in}
%\noindent
\textbf{Lighting normalization:} We design a \textit{novel} algorithm for removing spatial and temporal lighting variations from the 3D decomposition of the face by exploiting traits such as facial symmetry and albedo constancy of the skin. This lighting normalization removes another confounding factor that can otherwise affect the speech-to-lips mapping.

%\vspace{0.1in}
%\noindent
\textbf{Data-efficient learning:}
Our model employs an encoder-decoder architecture that computes embeddings from audio spectrograms, and decodes them to predict the decomposed 3D geometry and texture. Pose and lighting normalization allows us to train this model in a data-efficient manner. The model complexity is greatly reduced, since the network is not forced to disentangle unrelated head pose and lighting changes from speech, allowing it to synthesize high quality lip-sync results even from short training videos ($2$-$5$ minutes long). Lighting normalization allows training and inference illumination to be different, which obviates the need to train under multiple lighting scenarios.
The model predicts \textit{3D talking faces} instead of just a 2D image, even though it learns just from video, broadening its applicability. Finally, pose and lighting normalization can be applied in a backward fashion to align and match the appearance of the synthesized face with novel target videos. See~\FIGREF{fig_flow} for an overview of our approach.

\vspace{0.05in}
\noindent Our \textbf{key technical contributions} are:
\begin{itemize}[leftmargin=*]
\vspace{-0.1in}
\item A method to convert arbitrary talking head video footage into a normalized space that decouples 3D pose, geometry, texture, and lighting, thereby enabling data-efficient learning and versatile high-quality lip-sync synthesis for video and 3D applications.
\vspace{-0.05in}
\item A novel algorithm for normalizing facial lighting in video that exploits 3D decomposition and face-specific traits such as symmetry and skin albedo constancy.

\vspace{-0.05in}
\item To our best knowledge, this is the first attempt at disentangling pose and lighting from speech via data pre-normalization for personalized models. 

\vspace{-0.05in}
\item An easy-to-train auto-regressive texture prediction model for temporally smooth video synthesis.

\vspace{-0.05in}
\item Human ratings and objective metrics suggest that our method outperforms contemporary audio-driven video reenactment baselines in terms of realism, lip-sync and visual quality scores.

\end{itemize}

\section{Related Work}
\noindent\textbf{Audio-driven 3D Mesh Animation:} These methods generate 3D face models driven by input audio or text, but do not necessarily aim for photorealism. In~\cite{wang2011text}, the authors learn a Hidden Markov Model (HMM) to map Mel-frequency Cepstral Coefficients (MFCC) to PCA model parameters. 
Audio features are mapped to Jali~\cite{edwards2016jali} coefficients in~\cite{zhou2018visemenet}.
In~\cite{Karras}, the authors learn to regress to 3D vertices of a face model conditioned on input audio spectrograms and simultaneously disambiguate variations in facial expressions unexplained by audio. In \cite{hussen2020modality}, the authors learn to regress blendshapes of a 3D face using the combined audio-visual embedding from a deep network.
%Recently, the authors in VOCA~\cite{voca} presented a framework to 
VOCA~\cite{voca} pre-registers subject-specific 3D mesh models using FLAME~\cite{flame} and then learns (using hours of high quality 4D scans) an offset to that template based on incoming speech, represented with DeepSpeech~\cite{hannun2014deep} features. 

\vspace{0.1in}
\noindent\textbf{Audio-driven Video Synthesis:} These methods aim to generate visually plausible 2D talking head videos, conditioned on novel audio. In~\cite{chen2018lip}, an audio-visual correlation loss is used to match lip shapes to speech, while maintaining the identity of the target face. In~\cite{chen2019hierarchical}, a two-stage cascaded network is used to first predict 2D facial landmarks from audio, followed by target frame editing conditioned upon these landmarks. In~\cite{vougioukas2019end}, the authors leverage a temporal GAN for synthesizing video conditioned on audio and a reference frame. They improve it further in~\cite{ijcv19} via a specialized lip-sync discriminator. In contrast to our approach, the above methods fail to produce full-frame outputs; instead they generate normalized cropped faces, whose lips are animated based on input audio and a reference frame.

\par Among efforts on full-frame synthesis, Video Rewrite~\cite{bregler1997video} was a pioneering work. It represented speech with phonetic labels and used exemplar-based warping for mouth animation.
Speech2Vid~\cite{chung2017you} learns a joint embedding space for representing audio features and the target frame, and uses a shared decoder to transform the embedding into a synthesized frame. 
X2Face~\cite{x2face} learns to drive a target frame with the head pose and expression of another source video, and it can optionally be also driven by an audio to animate a target frame.
A framework to translate an input speech to another language and then modify the original video to match it is presented in~\cite{kr2019towards}.
Recently, Wav2Lip~\cite{wav2lip} reported appreciable lip-sync performance by using a powerful offline lip-sync discriminator~\cite{syncnet} as an expert to train their generator. While currently this is one of the best universal models, it lacks the visual fidelity of speaker-specific models.

\par Some recent works~\cite{obama, thies2020neural, linsen2020ebt} have focused on 3D model guided video synthesis. 
In~\cite{obama} an RNN regresses audio to mouth shape, producing convincing results on President Obama. The approach required very extensive training data however (17 hours).
In~\cite{thies2020neural}, the DeepSpeech RNN is used to map input speech to audio expression units which then drive a blendshapes-based 3D face model. Finally, a neural renderer~\cite{kim2018DeepVideo} is used to render the face model with the audio expressions. Since neural renderer training depends on target illumination, the methods leveraging such rendering \cite{linsen2020ebt, thies2020neural} suffer from the need for retraining if inference-time lighting conditions change. On the contrary, our method seamlessly adapts to novel lighting.

\vspace{0.1in}
\noindent\textbf{Text-based Video Editing:}
In~\cite{fried2019text}, the authors present a framework for text based editing of videos (TBE). They first align written transcripts to audio and track each frame to create a face model. During edit operations, a (slow) viseme search is done to find best matching part of training video. This method needs a time-aligned transcript and around one hour of recorded data, and is mostly suitable for for small edits. Our method, on the other hand, relies on just the audio signal and can synthesize videos of unrestricted length.

\vspace{0.1in}
\noindent\textbf{Actor-driven Video Synthesis:} \cite{face2face,kim2018DeepVideo} present techniques for generating and dubbing talking head videos by transferring facial features, such as landmarks or blendshape parameters, from a different actor's video. 
These techniques generate impressive results, however they require a video of a surrogate actor to drive synthesis. We emphasize that our approach uses only audio or text-to-speech (TTS) as the driving input, and does not require any actors for dubbing. 
It is therefore fundamentally different from these methods.
\begin{figure}[!t]
    \centering
    \includegraphics[width=\columnwidth]{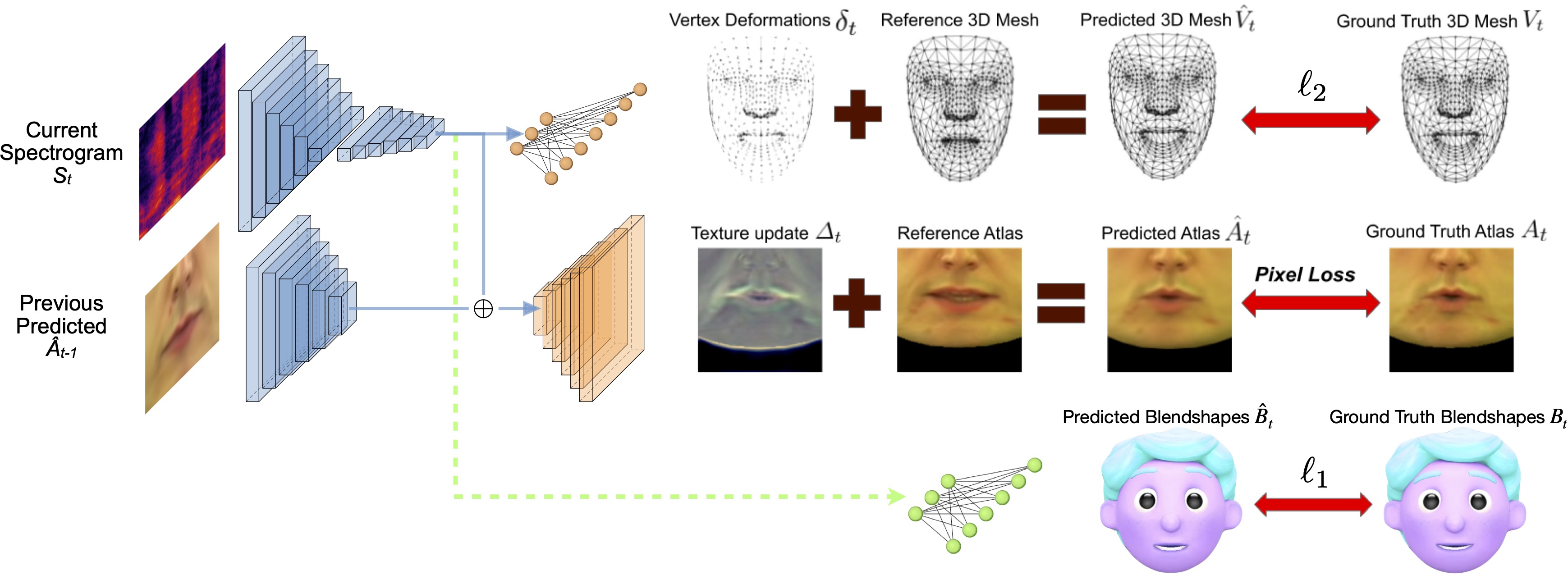}   % ARXIV version
\vspace{-0.4cm}
\caption{ \scriptsize Joint prediction pipeline: geometry and texture models have dedicated decoders but share the audio encoder. The texture model also depends on the previously predicted atlas. Optionally, the audio embedding can drive a 3D CGI character via a blendshape coefficients decoder. Please enlarge to see details.}  %was: Flow diagram of 
\label{fig_models} 
\vspace{-0.20cm}
\end{figure}

\section{Method}
\vspace{-0.5em}

We now describe the various components of our approach including data extraction and normalization, neural network architecture and training, and finally, inference and synthesis. Figure~\ref{fig_models} shows an overview of our model.

 We extract the audio channel from the training video and transform it into frequency-domain spectrograms. These spectrograms are computed using Short-time Fourier transforms (STFT) with a Hann window function~\cite{wikiSTFT}, over $30$ms wide sliding windows that are $10$ms apart. We align these STFTs with video frames and stack them across time to create a $256 \times 24$ complex spectrogram image, spanning $240$ms centered around each video frame. Our model predicts the face geometry, texture, and optionally, blendshape coefficients, for each frame based on the audio spectrogram.

The face in the video is tracked using a 3D face landmark detector \cite{xeno}, resulting in $468$ facial features, with the depth (z-component) predicted using a deep neural network. We refer to these features as vertices, which are accompanied by a predefined triangulated face mesh with fixed topology.

%============ Getting a 2-row diagram for CVPR'21 for space constraints
\begin{figure}
    \centering
    \includegraphics[width=\columnwidth]{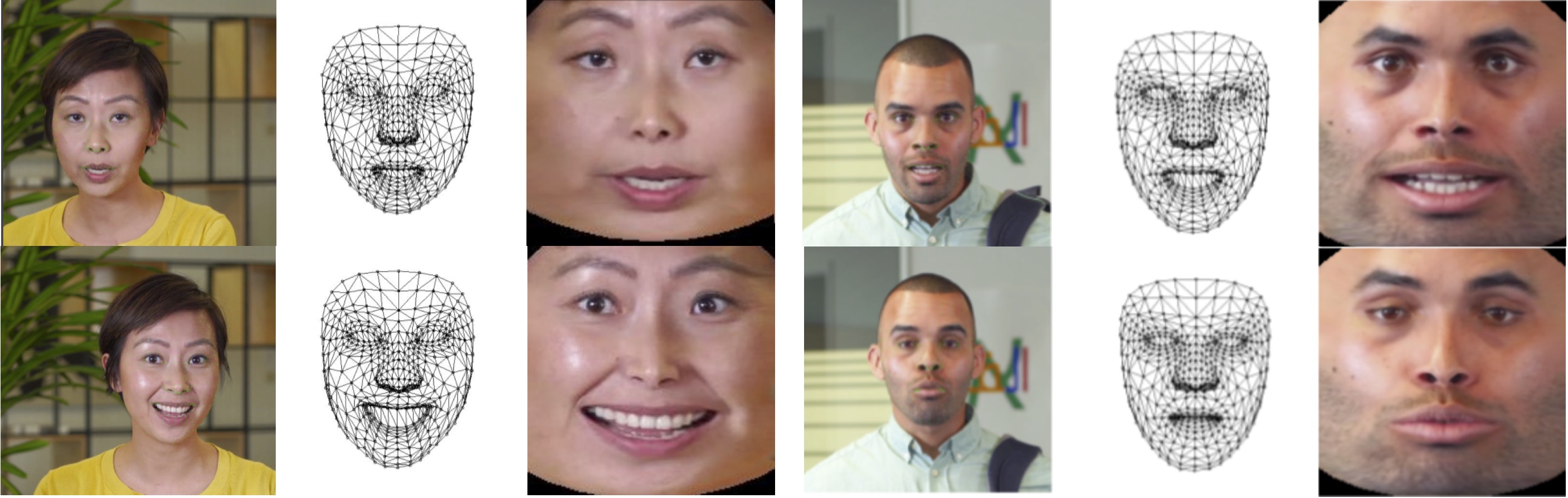}
    \caption{\scriptsize Pose normalization of training data. For each subject-- Left: input frames with detected features (see zoomed in); Middle: normalized vertices and triangle mesh; Right: texture atlas which acts as ground truth for texture prediction.}
    \label{fig:normalization}
\end{figure}

%=======Figure to show different steps of lighting normalization
\begin{figure*}
    \centering
    \includegraphics[scale=0.13]{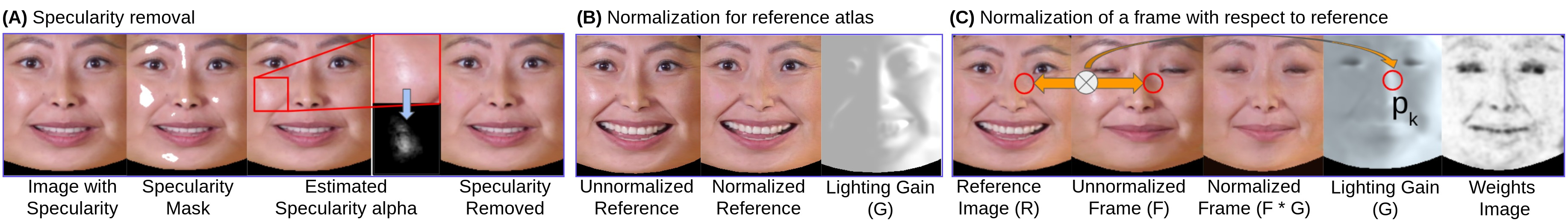}
    \caption{\scriptsize Steps of our proposed lighting normalization during training. \textbf{(A:)}~ First step is to specularity removal from an input frame. \textbf{(B:)}~ Second step is self normalization of the reference atlas. \textbf{(C:)}~ Finally, any given training frame is normalized with respect to the pre-normalized reference atlas of step B.}
    \label{fig_light_ablation}
\vspace{-1em}
\end{figure*}
%==================

\subsection{Normalizing Training Data} \label{normalizing}

We preprocess the training data to eliminate the effects of head movement and lighting variations, and work with normalized facial geometry and texture. Both training and inference take place in this normalized space.

\vspace{-0.3cm}
\subsubsection{Pose normalization}
For pose normalization, we first select one frame of the input video as a reference frame, and its respective 3D face feature points as reference vertices. The choice of frame is not critical; any frame where the face is sufficiently frontal is suitable. Using the reference vertices, we define a reference cylindrical coordinate system (similar to~\cite{booth2014uvspaces}) with a vertical axis such that most face vertices are equidistant to the axis. We then scale the face size such that the eyes and nose project to fixed locations on this reference cylinder.

Next, for each frame of the training video, we stabilize the rigid head motion (see \cite{disneyheadstab,EAheadstab}) to provide a registered 3D mesh suitable for training our geometry model. Specifically, we approximately align the vertices of the upper, more rigid parts of the face with corresponding vertices in the normalized reference using Umeyama's algorithm~\cite{Umeyama} and apply the estimated rotation $\RR$, translation $\btt$ and scale $c$ to all tracked vertices $\vv$ as $\hat{\rr} = c \RR \vv + \btt$.

We use these normalized vertices, along with the cylindrical mapping defined above, to create a pose-invariant, \textit{frontalized} projection of the face texture for each video frame (including the reference frame). Mapping the face vertices to the reference cylinder creates a set of 2D texture coordinates for the face's surface, which are used to \textit{unroll} its texture. We warp the triangles associated with these coordinates from the source frame onto the texture domain, resulting in a  $256\times 256$ \textit{texture atlas} that resembles a frontal view of the face, but with the non-rigid features like the lips and mouth moving with the speech. \FIGREF{fig:normalization} demonstrates the effect of normalization; the head pose is removed, but the moving lip shapes and mouth interior are preserved.

\vspace{-0.2cm}
\subsubsection{Lighting normalization}
%========================================================================
We normalize the frontalized texture atlas to remove lighting variations, which are mostly caused by head motion or changing illumination. Our lighting normalization algorithm works in two phases. It first exploits facial symmetry to normalize the reference atlas $R$ \textit{spatially}, removing specularities and lighting variations that run across the face. It then performs a \textit{temporal} normalization across video frames that transforms each frame's atlas $F$ to match the illumination of $R$. The resulting atlases have a more uniform \textit{albedo}-like appearance, that stays consistent across frames.

We first describe the temporal normalization algorithm, as it is a core component also used during spatial normalization. This algorithm assumes that the two textures $F$ and $R$ are pre-aligned geometrically. However, any non-rigid facial movements,~\eg from speech, can result in different texture coordinates, and consequently, misalignments between $R$ and $F$. Hence, we first warp $R$ to align it with $F$'s texture coordinates, employing the same triangle-based warping algorithm used for frontalization.

%Since $F$ and $R$ may have slightly different texture coordinates, especially in the lip regions, we first warp $R$ to align with $F$.

Given the aligned $R$ and $F$, we estimate a mapping that transforms $F$ to match the illumination of $R$. This mapping is  composed of a smooth multiplicative pixel-wise gain $G$ in the luminance domain, followed by a global channel-wise gain and bias mapping $\{a,b\}$ in the RGB domain. The resulting normalized texture $F^{n}$ is obtained via the following steps:
(1) $(F_y, F_u, F_v) = \textbf{\textsc{rgb}to\textsc{yuv}}(F)$;
(2) $F^{l}_y = G * F_y$;
(3) $F^{l} = \textbf{\textsc{yuv}to\textsc{rgb}}(F^{l}_y, F_u, F_v)$;
(4) $F^{n} = aF^{l} + b$.

%\vspace{0.1in}
%\noindent 
\textbf{Gain Estimation:} To estimate the gain $G$, we observe that a pair of corresponding pixels at the same location $k$ in $F$ and $R$ should have the same underlying appearance, modulo any change in illumination, since they are in geometric alignment (see~\FIGREF{fig_light_ablation}(C)). This \textit{albedo constancy} assumption, if perfectly satisfied, yields the gain at pixel $k$ as $G_k = R_k / F_k$. However, we note that (a) $G$ is a smoothly varying illumination map, and (b) albedo constancy may be occasionally violated,~\eg in non-skin pixels like the mouth, eyes and nostrils, or where the skin deforms sharply, \eg the nasolabial folds. We account for these factors by, firstly, estimating $G_k$ over a larger patch $p_k$ centered around $k$, and secondly, employing a robust estimator that weights pixels based on how well they satisfy albedo constancy. We formulate estimating $G_k$ as minimizing the error:
\begin{equation}
  \mathbf{E}_k = \sum_{j\in p_k} W_j  \|R_j - G_k * F_j\| ^ 2,
\end{equation}
where $W$ is the per-pixel weights image, and solve it using iteratively reweighted least squares (IRLS). In particular, we initialize the weights uniformly, and then update them after each ($i^{th}$) iteration as:
\begin{equation}
    W^{i+1}_k = \exp \left({\frac{-\mathbf{E}^i_k}{T}}\right),
%    W^{i+1}_k = \exp(-\mathbf{E}^i_k / T),
    %W^{i+1}_k = \exp \left( {\frac{-T.\mathbf{E}^i_k}{\sum_j{W^i_j}}}\right),
\end{equation}
where $T$ is a temperature parameter. The weights and gain converge in 5-10 iterations; we use $T=0.1$ and a patch size of $16\times 16$ pixels for $256\times 256$ atlases. \FIGREF{fig_light_ablation}(C) shows example weights and gain images.
Pixels with large error $\mathbf{E}_k$ get low weights, and implicitly interpolate their gain values from neighboring pixels with higher weights.

%We note that since the gain inside the mouth is most unreliable, we make use of face geometry to remove the weights in the mouth region.

To estimate the global color transform $\{a, b\}$ in closed form, we minimize $\sum_k W_k \| R_k - aF_k - b \|^2$ over all pixels, with $W_k$ now fixed to the weights estimated above.

\textbf{Reference Atlas Normalization using Facial Symmetry:}
We first estimate the gain $G^m$ between the reference $R$ and its mirror image $R'$, using the algorithm described above. 
This gain represents the illumination change between the left and right half of the face. % may be evident from previous sentence and context
To obtain a reference with uniform illumination, we compute the symmetrized gain $G^s = \max(G^m, {G^m}')$, where ${G^m}'$ is the mirror image of $G^m$, \ie for every symmetric pair of pixels, we make the darker pixel match the brighter one. The normalized reference is then $R^n = G^s * R$, as shown in~\FIGREF{fig_light_ablation}(B). Note that our weighting scheme makes the method robust to inherent asymmetries on the face, since any inconsistent pixel pairs will be down-weighted during gain estimation, thereby preserving those asymmetries.

%\vspace{0.1in}
%\noindent
\textbf{Specularity Removal:}
We remove specularities from the face before normalizing the reference and video frames, since they are not properly modeled as a multiplicative gain, and also lead to duplicate specularities on the reference due to symmetrization. We model specular image formation as:
\begin{equation}
  I = \alpha  + (1 - \alpha) * I_c,
\end{equation}
where $I$ is the observed image, $\alpha$ is the specular alpha map and $I_c$ is the underlying \textit{clean} image without specularities. We first compute a mask, where $\alpha > 0$, as pixels whose minimum value across RGB channels in a smoothed $I$ exceeds the $90^{th}$ percentile intensity across all skin pixels in $I$. The face mesh topology is used to identify and restrict computation to skin pixels. We then estimate a \textit{pseudo} clean image $\tilde{I}_c$ by hole-filling the masked pixels from neighboring pixels, and use it to estimate $\alpha = (I - \tilde{I}_c)/(1 - \tilde{I}_c)$. 
 
The final clean image is then $I_c = (I - \alpha) / (1 - \alpha)$. 
Note that our soft alpha computation elegantly handles any erroneous over-estimation of the specularity mask (see \FIGREF{fig_light_ablation}(A)).
The above method is specifically tailored for stabilized face textures and is simple and effective, thus we do not require more generalized specularity removal techniques~\cite{yang2014efficient}.

%==============================================================%
% Joint model description..

\newcommand{\Afix}{A^{\text{\emph{fix}}}}
\newcommand{\Alight}{A^{\text{\emph{light}}}}
\newcommand{\Aref}{A^{\text{\emph{ref}}}}
\newcommand{\mfun}{\mathbf{F}}
\newcommand{\Rgeo}{\mathbf{R_{geo}}}
\newcommand{\Rtex}{\mathbf{R_{tex}}}
\newcommand{\Rbs}{\mathbf{R_{bs}}}

\subsection{Joint Prediction Model and Training Pipeline}
\label{sec_model_pipeline}
In this section we describe the framework for learning a function $\mfun$ to jointly map from domain $S$ of audio spectrograms to the domains $V$ of \textit{vertices} and $A$ of texture \textit{atlases}: $\mfun:~S\rightarrow~V{\times}A$, with $V\in \mathbb{R}^{468\times 3}$ and $A\in \mathbb{R}^{128 \times 128 \times 3}$, where for the purpose of prediction, we crop the texture atlas to a $128 \times 128$ region around the lips, and only predict these cropped regions. The texture for the upper face is copied over from the reference, or target video frames, depending upon the application. We follow an encoder-decoder architecture for realizing $\mfun(\cdot)$, as shown in~\FIGREF{fig_models}. It consists of a shared encoder for audio, but separate dedicated decoders for geometry and texture. However, the entire model is trained jointly, end-to-end.
%Please see Appendix~\ref{appendix:arch} for more details on our network architecture.\\
%==============================================

% \vspace{0.1in}
\textbf{Audio encoder:} The input at time instant $t$ is a complex spectrogram,  $S_t\in \mathbb{R}^{256 \times 24 \times 2}$. Our audio encoder --- and face geometry prediction model --- is inspired by the one proposed in \cite{Karras}, in which the vertex positions of a fixed-topology face mesh are also modified according to an audio input. However, while~\cite{Karras} used formant preprocessing and autocorrelation layers as input, we directly use complex spectrograms $S_t$. Each $S_t$ tensor is passed through a $12$ layer deep encoder network, where the first $6$ layers apply 1D convolutions over frequencies (kernel $3 \times 1$, stride $2 \times 1$), and the subsequent $6$ layers apply 1D convolution over time (kernel $1 \times 3$, stride $1 \times 2$), all with leaky ReLU activation, intuitively corresponding to phoneme detection and activation, respectively. This yields a latent code $L_t^s \in \mathbb{R}^{N_s}$.

% \vspace{0.1in}

\textbf{Geometry decoder:~} This decoder maps the latent audio code $L_t^s$ to vertex \textit{deformations} $\delta_t$, which are added to the reference vertices $V_r$ to obtain the predicted mesh $\hat{V}_t = V_r + \delta_t$. It consists of two fully connected layers with $150$ and $1404$ units, and linear activations, with a dropout layer in the middle. The resulting output is  $468$ vertices ($1404 = 468 \times 3$ coordinates). As proposed in~\cite{Karras}, we initialize the last layer using PCA over the vertex training data. Further, we impose $\ell_2$ loss on the vertex positions: $\Rgeo = \|V_t - \hat{V}_t\|_2$, where $V_t$ are ground-truth vertices.

% \vspace{0.1in}
\textbf{Texture decoder:~} This decoder maps the audio code $L_t^s$ to a texture atlas \textit{update} (difference map) $\Delta_t$ which is added to the reference atlas $A_{r}$ to obtain the predicted atlas, $\hat{A}_t = A_r + \Delta_t$. It consists of a fully connected layer to distribute the latent code spatially, followed by progressive up-sampling using convolutional and interpolation layers to generate the $128\times128$ texture update image (see Appendix \ref{sec_texture_decoder}).  We impose an image similarity loss between the predicted and ground-truth atlas $A_t$: $\Rtex = d(A_t, \hat{A}_t)$, where $d$ is a visual distance measure. We tried different variants of $d(\cdot)$ including the $\ell_1$ loss, Structural Similarity Loss (SSIM), and Gradient Difference Loss (GDL)~\cite{mathieu2015deep} and found SSIM to perform the best.

\textbf{Blendshapes decoder:~} To animate CGI characters using audio, we optionally add another decoder to our network that predicts \textit{blendshape} coefficients $B_t$ in addition to geometry and texture. For training, these blendshapes are derived from vertices $V_t$ by fitting them to an existing blendshapes basis either via optimization or using a pre-trained model \cite{blendshapesurvey14}. We use a single fully connected layer to predict coefficients $\hat{B}_t$ from audio code $L_t^s$, and train it using $\ell_1$ loss $\Rbs = \|B_t - \hat{B}_t\|_1$ to encourage sparse coefficients.

\vspace{-1em}
\subsubsection{Auto-regressive (AR) Texture Synthesis:}
\label{sec_ar}

Ambiguities in facial expressions while speaking (or silent) can result in temporal jitters. We mitigate these by incorporating memory into the network. 
Rather than using RNNs, we condition the current output of the network ($A_t$) 
not only on $S_t$ but also on the previous predicted atlas $\hat{A}_{t-1}$, encoding it as a latent code vector $L_{t-1}^a \in \mathbb{R}^{N_a}$. 
$L_t^s$ and $L_{t-1}^a$ are combined and passed to the texture decoder to generate the current texture $\hat{A}_t$ (Figure \ref{fig_models}). This appreciably improves the temporal consistency of synthesized results. We can train this AR network satisfactorily via \textit{Teacher Forcing}~\cite{williams:recurrent}, using previous ground truth atlases.
The resulting network $\mfun$ is trained end-to-end, minimizing the combined loss $\mathbf{R} = \Rtex + \alpha_1 \Rgeo + \alpha_2 \Rbs$, where $\alpha_1=3.0$ and $\alpha_2=0.3$ (when enabled). We used hyperparameter search to determine the latent code lengths, $N_s=32$ and $N_a=2$.
% The two figures from ablation studies
% moved to a separate file to allow them to be more easily moved
% while trying to compress space

%=========Figure to show benefit of AR Modeling
\begin{figure*}[th]
    \centering
    \includegraphics[width=0.92\textwidth]{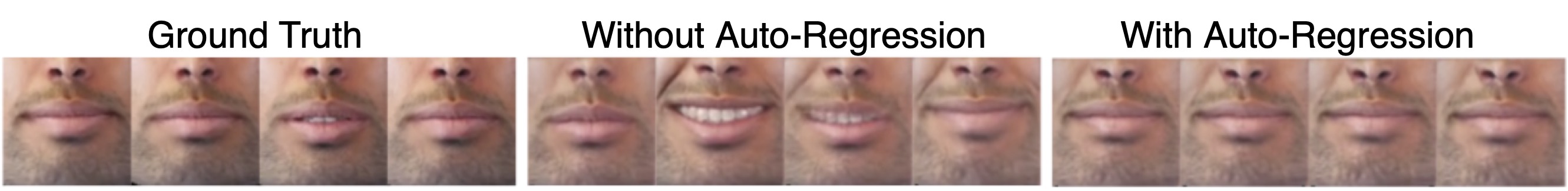}
    \caption{\scriptsize Benefits of proposed auto-regressive (AR) prediction. Left: Four consecutive frames when the subject was silent. Middle: Prediction without AR. Right: Prediction with AR. In absence of AR, the model fluctuates between different visual states, while the AR substantially improves temporal stability.}
    \label{fig_AR_model}
 \vspace{-1em}
\end{figure*}

%================ Figure to show benefit of lighting normalization
% \begin{figure*}
%     \centering
%     \includegraphics[scale=0.17]{ImagesCVPR2021/light.jpg}
%     \caption{Importance of proposed lighting normalization. \textbf{Row 1:}~The model is trained on a short indoors video in which the lighting is almost constant. Inference is performed under a varying illumination setting. Without lighting normalization, we see distinct artefacts near the lip region since the texture generation network is unable to match the novel and changing lighting condition. \textbf{Row 2:}~The model is trained outdoors under a harsh and varying lighting. In this tougher training scenario, the model without lighting normalization degrades further. Under both of these cases, proposed model is able to generate visually plausible texture with appropriate lip-sync. Please zoom in for better visualization. Videos are provided in supplementary.}
%     \label{fig_light_normalization}
% \end{figure*}
\begin{figure}
    \centering
    \includegraphics[scale=0.12]{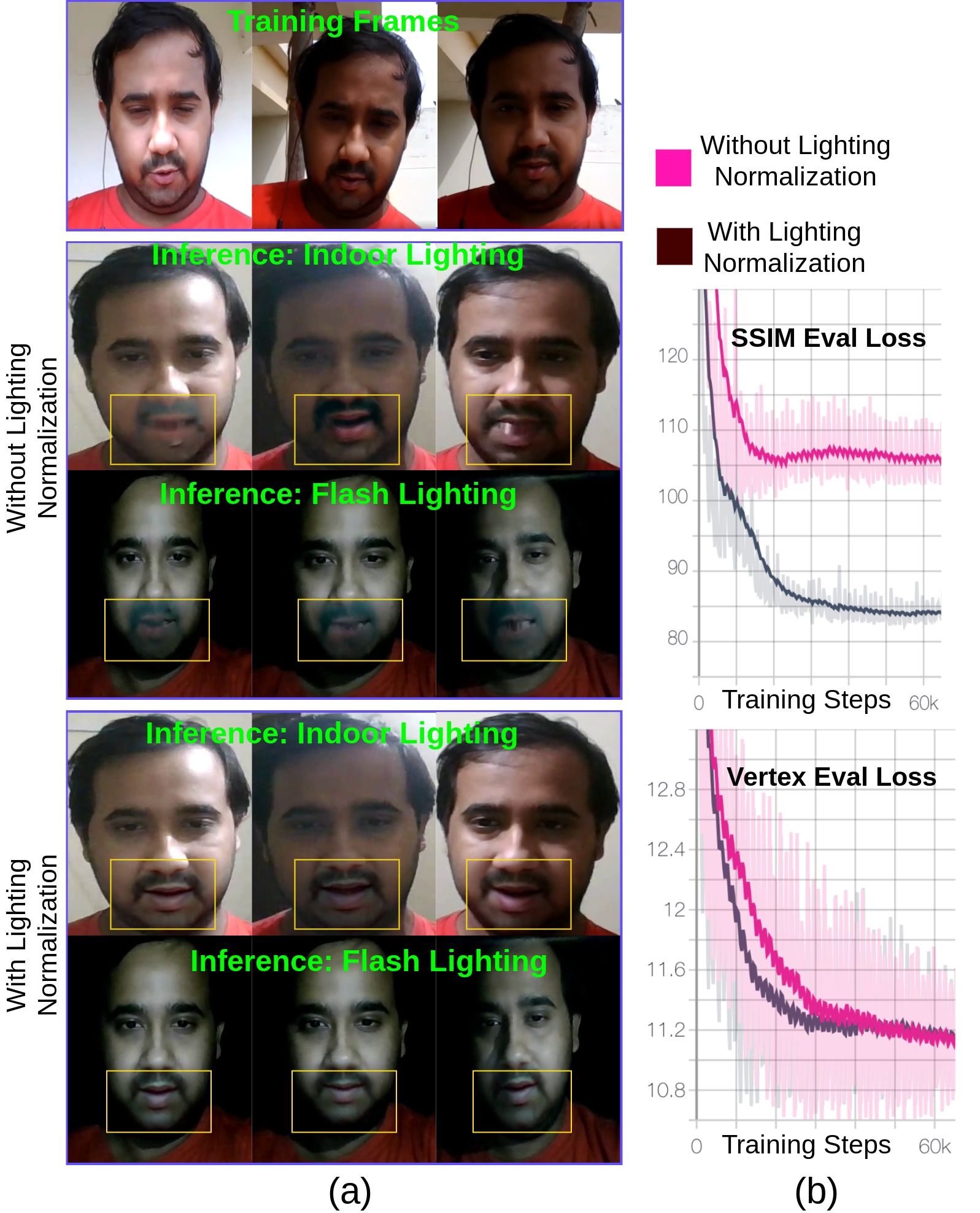}
    \caption{\scriptsize \textbf{(a:)}~Benefits of the proposed lighting normalization. Top row shows representative training frames in a sunny outdoor setting while we conduct inference under two novel lighting settings which have not been used in training. Note that the proposed lighting normalization enables realistic synthesis under new lighting while absence of lighting normalization yields degraded outputs.~\textbf{(b:)}~ Plot of SSIM loss (texture prediction) and vertex loss (geometry prediction) on the evaluation set. Even though both models result in similar lip shapes, the lower SSIM loss of the lighting-normalized model boosts the visual realism and overall lip-sync quality.}
    \label{fig_light_normalization}
\vspace{-0.4cm} %jp/space
\end{figure}

\subsection{Inference and Synthesis}

%\noindent.
\textbf{Textured 3D mesh:}
During inference, our model predicts geometry and texture from audio input. To convert it to a textured 3D mesh, we project the predicted vertices onto the reference cylinder, and use the resulting 2D locations as texture coordinates. Since our predicted texture atlas is defined on the same cylindrical domain, it is consistent with the computed texture coordinates. The result is a fully textured 3D face mesh, driven by audio input (\FIGREF{fig_flow_mesh}).
%================

%\vspace{0.1in}
%\noindent 
\textbf{Talking head video synthesis:~}\label{insert_video}
The pose and lighting normalization transforms 
(\SECREF{normalizing}) are invertible,~\ie one can render the synthesized face mesh in a different pose under novel lighting, which allows us to procedurally blend it back into a different target video (\FIGREF{fig_flow_video}).
Specifically, we warp the textured face mesh to align it with the target face, then apply our lighting normalization algorithm in reverse,~\ie on the warped texture, using the target face as reference. One caveat is that the target frame's area below the chin may not align with the warped synthesized face, due to inconsistent non-rigid deformations of the jaw. Hence, we pre-process each target frame by warping the area below the original chin to match the expected new chin position. To avoid seams at border areas,
we gradually blend
between the original and new face geometry, and warp the original face in the target frame according to the blended geometry. 

\textbf{Cartoon rendering:} For stylized visualizations, we can create a cartoon rendering of the textured mesh (or video), by combining bilateral filtering with a line drawing of the facial features. In particular, we identify nose, lips, cheeks and chin contours in the synthesized face mesh, and draw them prominently over the filtered texture or video frame.

 \textbf{CGI Characters:} Models trained with the blendshapes decoder also output blendshape coefficients that can drive a CGI character. We combine these predicted blendshapes (that generally affect the lips and mouth) with other blendshapes, such as those controlling head motion and eye gaze, to create lively real-time animations.
 Please refer to Appendix- \ref{sec_supple_applications} and Fig. \ref{fig_applications} for more details.

%================== Begin section on Experiments.
\section{Experiments}
Our training and inference pipelines were implemented in Tensorflow~\cite{tensorflow2015-whitepaper}, Python and C++. We trained our models with batch sizes of $128$ frames,  for $500$-$1000$ epochs, with each epoch spanning the entire training video. Sample training times were between $3$-$5$ hours, depending on video length (usually $2$-$5$min). Average inference times were $3.5$ms for vertices, $31$ms for texture and $2$ms for blendshapes, as measured on a GeForce GTX 1080 GPU. Our research-quality code for blending into target videos takes $50$-$150$ms per frame, depending on the output resolution.
%=========== Figure to compare on GRID, TCD and CREMA
\begin{figure*}
    \centering
    \includegraphics[width=0.92\linewidth]{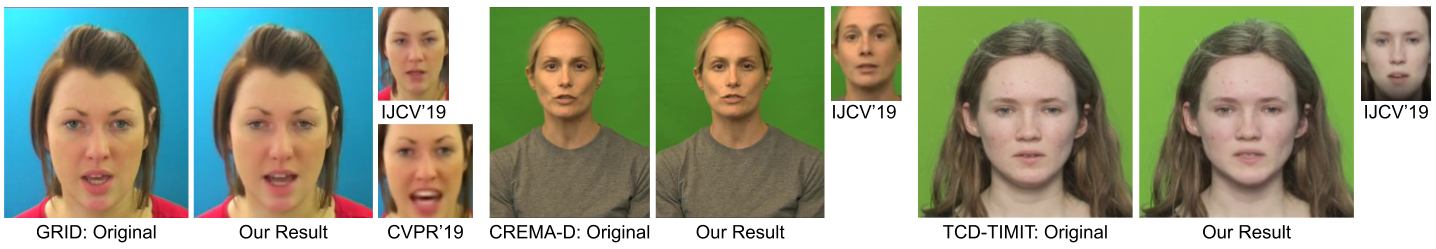}
    \caption{\scriptsize Qualitative comparison on subjects from GRID, CREMA-D and TCD-TIMIT against IJCV'19 and CVPR'19 (latter only available on GRID). 
    %Our result better matches the lip shape and resolution of the ground truth. 
    Our model is capable of seamlessly blending back into the video instead of animating a normalized cropped frame as in IJCV'19 and CVPR'19..}
    \label{fig_compare_all}
\end{figure*}

%=============== Figure to compare with Wav2Lip
\begin{figure}
\vspace{-0.3cm}
    \centering
    \includegraphics[scale=0.16]{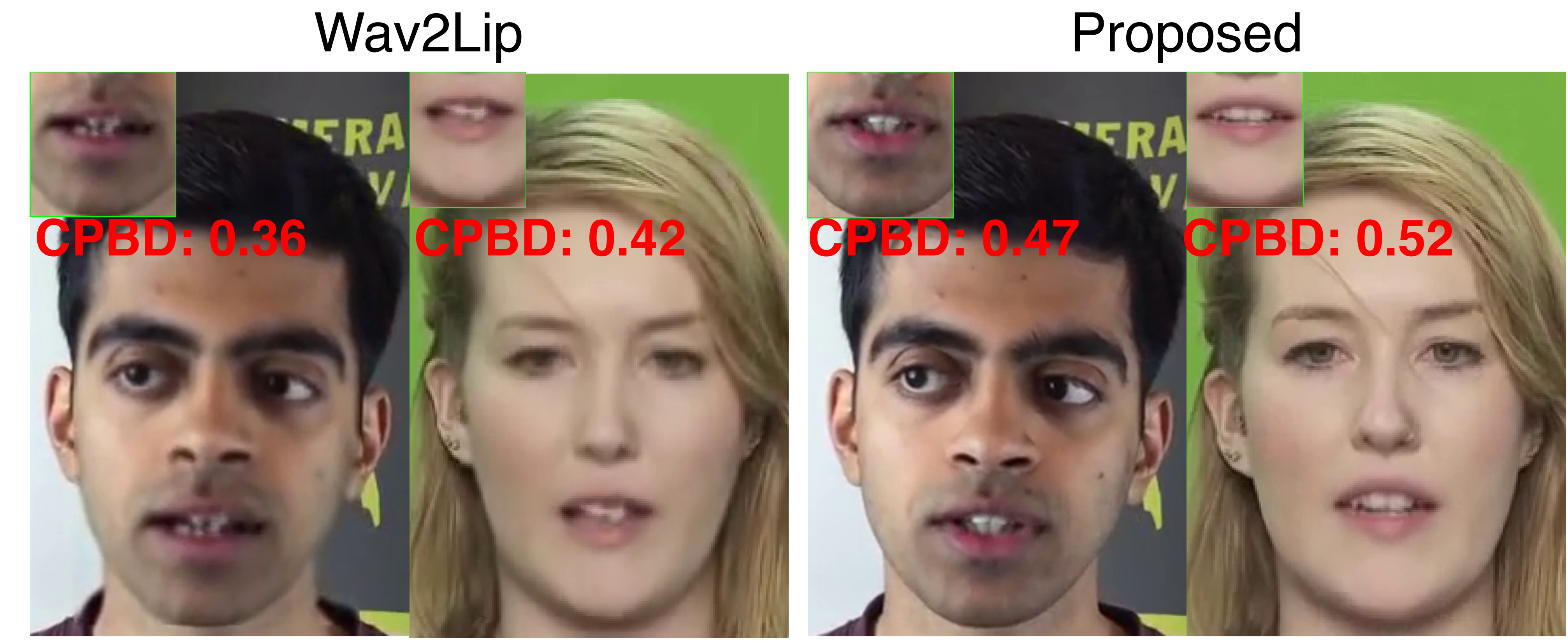}
    \caption{\scriptsize Comparison with Wav2Lip~\cite{wav2lip}. Our model generates higher resolution outputs (evident by higher CPBD metric~\cite{cpbd}) with fewer artifacts compared to Wav2Lip. Examples are provided in accompanying video.}
    \label{fig_wav2lip}
\end{figure}

\subsection{Ablation Studies}

%\vspace{0.1in}
\textbf{Benefit of Auto-Regressive Prediction:}
%We observed that 
The auto-regressive texture prediction algorithm stabilizes mouth dynamics considerably. In~\FIGREF{fig_AR_model}, we show that without auto-regression, the model can produce an unrealistic jittering effect, especially during silent periods. 
%======================

\textbf{Benefit of Lighting Normalization:}
%Here, we illustrate the importance of lighting normalization.
We use a short training video ($\sim$4 minutes) recorded in an outdoor setting but with varying illumination. However, during inference, we select two novel environments: a)~indoor lighting with continuous change of lighting direction, and b)~a dark room with a face illuminated by a moving flash light. Some representative frames of models trained with and without lighting normalization are shown in \FIGREF{fig_light_normalization}(a). Without lighting normalization, the model produces disturbing artifacts around the lip region, exacerbated by the extreme changes in illumination. However, with normalized lighting, the model adapts to widely varying novel illumination conditions.
This ability to edit novel videos of the same speaker \textit{on-the-fly} without needing to retrain for new target illumination is a significant benefit. In contrast, neural rendering based approaches~\cite{thies2020neural} require retraining on each new video, because they map 3D face models directly to the facial texture in video without disentangling illumination.

We also visualize the loss curves on held out evaluation sets in \FIGREF{fig_light_normalization}(b). %For the model trained 
With lighting normalization, the SSIM loss (used for texture generation) saturates at a much lower value than without normalization. This supports our hypothesis that lighting normalization results in more data-efficient learning, since it achieves a better loss with the same amount of training data. The vertex loss (responsible for lip dynamics) is similar for both models, because lighting normalization does not directly affect the geometry decoder, but overall lip-sync and visual quality  are improved.

\subsection{Comparison: Self-reenactment}
We objectively evaluate our model under the self-reenactment setting (audio same as target video), since it allows us to have access to ground truth facial information. We show experiments with three talking head datasets: GRID~\cite{Cooke2006}, TCD-TIMIT~\cite{tcd} and CREMA-D~\cite{crema}.

%================================
\begin{figure}
    \centering
    \includegraphics[scale=0.15]{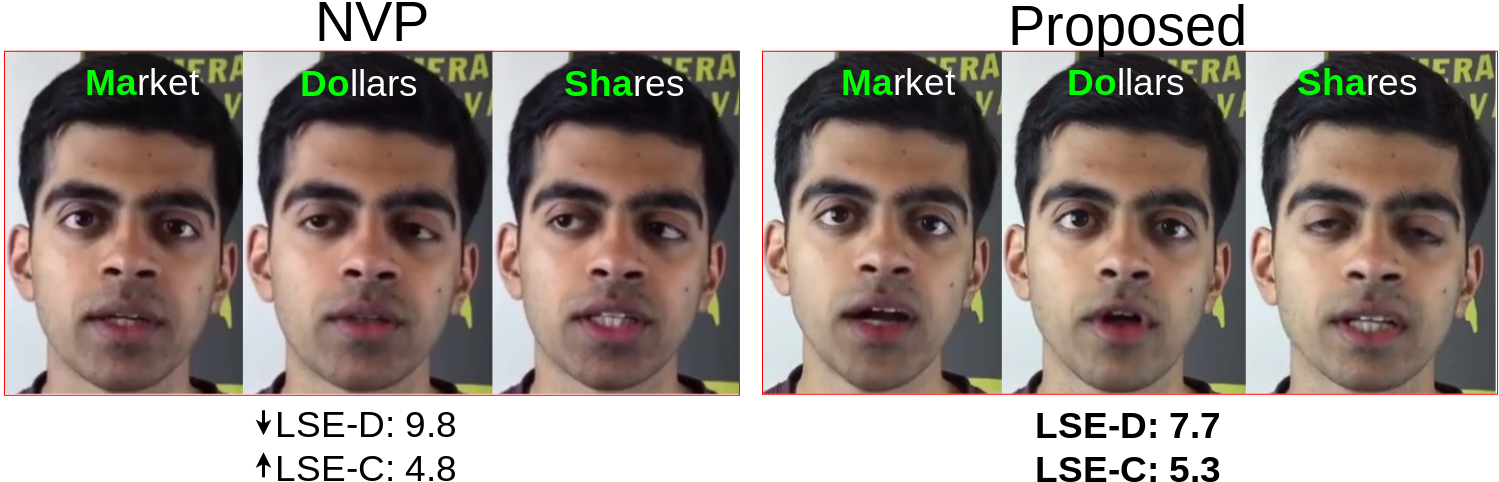}
    \caption{\scriptsize Comparison with NVP~\cite{thies2020neural}. We show that a sequence generated by our method usually has better lip dynamics compared to NVP. The observation is also supported by LSE-D (lower is better) and LSE-C (higher is better) metrics~\cite{wav2lip} for our model. Examples are provided in accompanying video. Best viewed \textbf{zoomed} in.}
    \label{fig_voca_nvp_compare}
 \vspace{-1em} %jp/space
\end{figure}
%==========Table to compare Self Reenactment + MOS
\begin{figure*}
    \centering
    \includegraphics[scale=0.24]{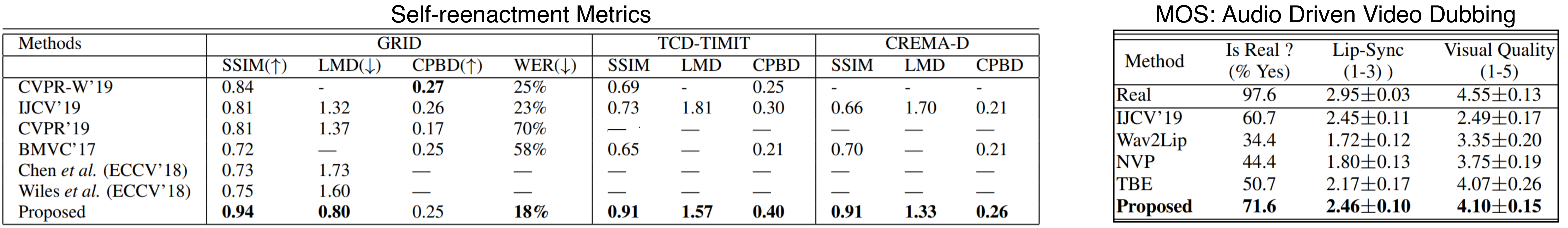}
    \caption{\scriptsize \textbf{Left:~} Self-reenactment performance comparison against state-of-the-art benchmarks of CVPR-W'19~\cite{vougioukas2019end}, IJCV'19~\cite{ijcv19}, CVPR'19~\cite{chen2019hierarchical}, BMVC'17~\cite{chung2017you}, Chen \textit{et al.}~\cite{chen2018lip} and  Wiles \textit{et al.}~\cite{x2face}. Pre-trained LipNet (for WER) is available only on GRID. Authors of \cite{chen2019hierarchical} released checkpoint for GRID only. ($\uparrow$):Higher is better. ($\downarrow$):Lower is better. Best results are marked in bold.~\textbf{Right:~} Mean Opinion Scores of user study. The statistical significance of these differences in ratings is confirmed by ANOVA with Tukey post-hoc tests. Please see Appendix- \ref{sec_user_study_analysis} for details. }
    \label{fig_combined_result}
\vspace{-0.4cm}  % Bad!  
\end{figure*}

%===========================

\textbf{Comparing Methods:} We perform quantitative comparisons against state-of-the-art methods whose models/results are publicly available: CVPR'19~\cite{chen2019hierarchical}, IJCV'19~\cite{ijcv19}, CVPR-W~\cite{vougioukas2019end}. 
It is difficult to do an apples-to-apples comparison, since we use personalized models while other techniques use a universal model. However, we minimize this gap by testing on the same 10 subjects from each of the 3 datasets used in IJCV'19 and CVPR'19, and employing the official evaluation frameworks of these papers. We also compare against other prior methods, but reuse the results already reported by CVPR'19 or IJCV'19.
Details of subject IDs are provided in Appendix- \ref{sec_subject_details}

\textbf{Evaluation Metrics:}
We follow the trend in recent papers \cite{chen2019hierarchical,ijcv19,vougioukas2019end}, which use \textbf{SSIM} (Structural Similarity Index) as a reconstruction metric, \textbf{LMD} (Landmark Distance) on mouth features as a shape similarity metric, \textbf{CPBD} (Cumulative Probability Blur Detection)~\cite{cpbd} as a sharpness metric and \textbf{WER} (word error rate) as a content metric to evaluate the correctness of words from reconstructed videos. Following \cite{ijcv19}, we use a LipNet model~\cite{assael2016lipnet} pre-trained for lip-reading on GRID dataset~\cite{Cooke2006}.
%============================

\textbf{Observations:}
We report the metrics in~\FIGREF{fig_combined_result} (left). 
On LMD and WER, which capture lip-sync, our model is significantly better than any competing method. Also, in terms of reconstruction measures (SSIM, CPBD), our model almost always performs better. 
CVPR-W and IJCV'19 have a better (though comparable) CPBD on GRID, but it is a low-resolution dataset.
On higher resolution TCD-TIMIT and CREMA-D, our CPBD is the best. 
We also show qualitative comparisons in~\FIGREF{fig_compare_all}. Note that we synthesize full frame videos, while CVPR'19 and IJCV'19 only generated normalized face crops at a resolution of $128 \times 128$, and $96 \times 128$ respectively. Thus our method is more suitable for practical video applications.
%==================================
\subsection{ Comparison: Audio-Driven Video Dubbing}
In this section we focus on \textit{`audio-driven'} video dubbing where the driving audio is different from the target video.

\textbf{User Study:} We conducted a user study to quantitatively compare our lip-sync and perceptual quality against the state-of-the-art audio-driven frameworks of Wav2Lip, NVP, IJCV'19 and TBE.  
In the study, 35 raters were each shown 29 sample clips consisting of synthetic and real videos. For competing methods, we used their released videos or generated results with their pre-trained models. The raters were asked three questions: Q1) Is the video real or fake? Q2) Rate lip-sync quality on a 3-point discrete scale. Q3) Rate visual quality on a 5-point discrete scale. We report the Mean Opinion Scores (MOS) of the questions in~\FIGREF{fig_combined_result} (right). As is evident, among the competing methods our method receives the most favorable user ratings.

%==========================
% \textbf{Measuring Lip-Sync Error:~} During video editing, in absence of ground truth, recently, authors in \cite{wav2lip} proposed two metrics, \textit{`Lip-Sync Error Distance (LSE-D)'} and \textit{`Lip-Sync Error Confidence (LSE-C)'} based on audio and video embedding coming from a pre-trained SyncNet~\cite{syncnet}. A lower LSE-D is preferred which means lip movements are in sync with audio. A higher LSE-C is recommended which refers to better audio-visual correlation.\\ 
%A lower LSE-C means there are multiple segments in the video where the lip movements are totally out of sync with audio \cite{syncnet}.\\
%==========================

\textbf{Comparison with Wav2Lip~\cite{wav2lip}:~} 
Unlike other image-based methods, %~\eg IJCV'19 and CVPR'19, 
%These methods, such as Wav2Lip, IJCV'19, CVPR'19 operate direcly on 2D image space. As discussed above, methods of IJCV'19 and CVPR'19 can only animate a cropped and normalized face-- thereby not feasible to realistic video editing.
Wav2Lip can paste back the generated face on background video. However, compared to our model, the outputs from Wav2Lip are of low resolution. Also, at high resolution, Wav2Lip produces significant visual artifacts (see~\FIGREF{fig_wav2lip}) and lip-sync starts to degrade.
%=============================

\textbf{Comparison with NVP~\cite{thies2020neural}:~} The lip-sync and dynamics of our model are generally better than NVP. 
 The lip movements of NVP are clearly muted compared to our model, as seen in representative frames in~\FIGREF{fig_voca_nvp_compare}(a).
 \vspace{-0.2cm}
\section{Applications}

\textbf{Speech/Text-to-Video:}  We can create or edit talking head videos for education, advertisement, and entertainment by simply providing new audio transcripts.
\textbf{ ``Actor-free'' video translation:} while \textit{`actor-driven'} video translation techniques~\cite{kim2018DeepVideo,face2face} generally require a professional actor to record the entire translated audio and video, our \textit{`actor-free'} approach  does not need video, and can be driven by either recorded audio, TTS, or voice cloning ~\cite{hsu2019_voicecloning}.
\textbf{Voice controlled Avatars:} Our model's blendshapes output can be used to animate CGI characters in real-time, allowing low-bandwidth voice-driven avatars for chat, VR, and games without the need for auxiliary cameras.
\textbf{Assistive technologies:} Voice-driven 3D faces can support accessibility and educational applications,~\eg personified assistants and cartoon animations for visualizing pronunciation.
\section{Limitations and Conclusion}

\textbf{Facial expressions:} We do not explicitly handle facial expressions, though our model may implicitly capture correlations between expressions and emotion in the audio track. \textbf{Strong movements in the target video:} When synthesized faces are blended back into a target video, emphatic hand or head movement might seem out of place. This has not proved to be a problem in our experiments. \textbf{Processing speed:} Our research-quality code, running at highest quality, is slightly slower than real-time.  
We have presented a data efficient yet robust end-to-end system for synthesizing personalized 3D talking faces, 
with applications in video creation and editing, 3D gaming and CGI. Our proposed pose and lighting normalization decouples non-essential factors such as head pose and illumination from speech and enables training our model on a relatively short video of a single person while nevertheless generating high quality lip-sync videos under novel ambient lighting. 
We envision that our framework is a promising stepping stone towards personalized audio-visual avatars and AI-assisted video content creation.

\section{Ethical Considerations}
Our technology focuses on world-positive use cases and applications.
Video translation and dubbing have a variety of beneficial and impactful uses, including making educational lectures, video-blogs, public discourse, and entertainment media accessible to people speaking different languages, and creating personable virtual ``assistants" that interact with humans more naturally.

However, we acknowledge the potential for misuse, especially since audiovisual media are often treated as veracious information. We strongly believe that the development of such generative models by \textit{good actors} is crucial for enabling preemptive research on fake content detection and forensics, which would allow them to make early advances and stay ahead of actual malicious attacks. %~\cite{?}. 
Approaches like ours can also be used to generate counterfactuals for training provenance and digital watermarking techniques.
We also emphasize the importance of acting responsibly and taking ownership of  synthesized content. To that end, we strive to take special care when sharing videos or other material that have been synthesized or modified using these techniques, by clearly indicating the nature and intent of the edits. Finally, we also believe it is imperative to obtain consent from all performers whose videos are being modified, and be thoughtful and ethical about the content being generated. We follow these guiding principles in our work.
\vspace{-0.50mm}
\section{Acknowledgments}
We would like to thank all the performers who graciously allowed us to use their videos for this work, including D.~Sculley, Kate Lane, Ed Moreno, Glenn Davis, Martin Aguinis, Caile Collins, Laurence Moroney, Paige Bailey, Ayush Tewari and Yoshua Bengio. We also thank the performers and creators of external, public datasets, as well as all participants of our user study. We would also like to thank our collaborators at Google and DeepMind: Paul McCartney, Brian Colonna, Michael Nechyba, Avneesh Sud, Zachary Gleicher, Miaosen Wang, Yi Yang, Yannis Assael, Brendan Shillingford, and Yu Zhang.
\bibliographystyle{cvprstyle/ieee_fullname}
\bibliography{egbib}

\begin{thebibliography}{10}\itemsep=-1pt

\bibitem{tensorflow2015-whitepaper}
Mart\'{\i}n Abadi, Ashish Agarwal, Paul Barham, Eugene Brevdo, Zhifeng Chen,
  Craig Citro, Greg~S. Corrado, Andy Davis, Jeffrey Dean, Matthieu Devin,
  Sanjay Ghemawat, Ian Goodfellow, Andrew Harp, Geoffrey Irving, Michael Isard,
  Yangqing Jia, Rafal Jozefowicz, Lukasz Kaiser, Manjunath Kudlur, Josh
  Levenberg, Dan Man\'{e}, Rajat Monga, Sherry Moore, Derek Murray, Chris Olah,
  Mike Schuster, Jonathon Shlens, Benoit Steiner, Ilya Sutskever, Kunal Talwar,
  Paul Tucker, Vincent Vanhoucke, Vijay Vasudevan, Fernanda Vi\'{e}gas, Oriol
  Vinyals, Pete Warden, Martin Wattenberg, Martin Wicke, Yuan Yu, and Xiaoqiang
  Zheng.
\newblock {TensorFlow}: Large-scale machine learning on heterogeneous systems,
  2015.
\newblock Software available from tensorflow.org.

\bibitem{assael2016lipnet}
Yannis~M Assael, Brendan Shillingford, Shimon Whiteson, and Nando De~Freitas.
\newblock Lipnet: End-to-end sentence-level lipreading.
\newblock {\em arXiv preprint arXiv:1611.01599}, 2016.

\bibitem{disneyheadstab}
Thabo Beeler and Derek Bradley.
\newblock Rigid stabilization of facial expressions.
\newblock {\em ACM Trans. Graph.}, 33(4), July 2014.

\bibitem{booth2014uvspaces}
J. {Booth} and S. {Zafeiriou}.
\newblock Optimal {UV} spaces for facial morphable model construction.
\newblock In {\em IEEE ICIP}, pages 4672--4676, 2014.

\bibitem{bregler1997video}
Christoph Bregler, Michele Covell, and Malcolm Slaney.
\newblock {V}ideo {R}ewrite: driving visual speech with audio.
\newblock In {\em SIGGRAPH}, volume~97, pages 353--360, 1997.

\bibitem{chen2018lip}
Lele Chen, Zhiheng Li, Ross K~Maddox, Zhiyao Duan, and Chenliang Xu.
\newblock Lip movements generation at a glance.
\newblock In {\em ECCV}, pages 520--535, 2018.

\bibitem{chen2019hierarchical}
Lele Chen, Ross~K Maddox, Zhiyao Duan, and Chenliang Xu.
\newblock Hierarchical cross-modal talking face generation with dynamic
  pixel-wise loss.
\newblock In {\em CVPR}, pages 7832--7841, 2019.

\bibitem{chung2017you}
Joon~Son Chung, Amir Jamaludin, and Andrew Zisserman.
\newblock You said that?
\newblock In {\em BMVC}, 2017.

\bibitem{syncnet}
J.~S. Chung and A. Zisserman.
\newblock Out of time: automated lip sync in the wild.
\newblock In {\em Workshop on Multi-view Lip-reading, ACCV}, 2016.

\bibitem{Cooke2006}
Martin Cooke, Jon Barker, Stuart Cunningham, and Xu Shao.
\newblock An audio-visual corpus for speech perception and automatic speech
  recognition.
\newblock {\em The Journal of the Acoustical Society of America},
  120(5):2421--2424, November 2006.

\bibitem{cudeiro2019capture}
Daniel Cudeiro, Timo Bolkart, Cassidy Laidlaw, Anurag Ranjan, and Michael~J
  Black.
\newblock Capture, learning, and synthesis of {3D} speaking styles.
\newblock In {\em CVPR}, pages 10101--10111, 2019.

\bibitem{voca}
Daniel Cudeiro, Timo Bolkart, Cassidy Laidlaw, Anurag Ranjan, and Michael~J
  Black.
\newblock Capture, learning, and synthesis of 3d speaking styles.
\newblock In {\em CVPR}, pages 10101--10111, 2019.

\bibitem{edwards2016jali}
Pif Edwards, Chris Landreth, Eugene Fiume, and Karan Singh.
\newblock {JALI:} an animator-centric viseme model for expressive lip
  synchronization.
\newblock {\em ACM TOG}, 35(4):127, 2016.

\bibitem{fried2019text}
Ohad Fried, Ayush Tewari, Michael Zollh{\"o}fer, Adam Finkelstein, Eli
  Shechtman, Dan~B Goldman, Kyle Genova, Zeyu Jin, Christian Theobalt, and
  Maneesh Agrawala.
\newblock Text-based editing of talking-head video.
\newblock {\em ACM TOG}, 38(4):1--14, 2019.

\bibitem{hannun2014deep}
Awni Hannun, Carl Case, Jared Casper, Bryan Catanzaro, Greg Diamos, Erich
  Elsen, Ryan Prenger, Sanjeev Satheesh, Shubho Sengupta, Adam Coates, et~al.
\newblock Deep speech: Scaling up end-to-end speech recognition.
\newblock {\em arXiv preprint arXiv:1412.5567}, 2014.

\bibitem{tcd}
Naomi Harte and Eoin Gillen.
\newblock {TCD-TIMIT}: An audio-visual corpus of continuous speech.
\newblock {\em IEEE Transactions on Multimedia}, 17(5):603--615, 2015.

\bibitem{hsu2019_voicecloning}
Wei-Ning Hsu, Yu Zhang, Ron Weiss, Heiga Zen, Yonghui Wu, Yuxuan Wang, Yuan
  Cao, Ye Jia, Zhifeng Chen, Jonathan Shen, Patrick Nguyen, and Ruoming Pang.
\newblock Hierarchical generative modeling for controllable speech synthesis.
\newblock In {\em International Conference on Learning Representations}, 2019.

\bibitem{hussen2020modality}
Ahmed Hussen~Abdelaziz, Barry-John Theobald, Paul Dixon, Reinhard Knothe,
  Nicholas Apostoloff, and Sachin Kajareker.
\newblock Modality dropout for improved performance-driven talking faces.
\newblock In {\em International Conference on Multimodal Interaction}, pages
  378--386, 2020.

\bibitem{Karras}
Tero Karras, Timo Aila, Samuli Laine, Antti Herva, and Jaakko Lehtinen.
\newblock Audio-driven facial animation by joint end-to-end learning of pose
  and emotion.
\newblock {\em ACM TOG}, 36(4):94:1--94:12, July 2017.

\bibitem{xeno}
Yury Kartynnik, Artsiom Ablavatski, Ivan Grishchenko, and Matthias Grundmann.
\newblock Real-time facial surface geometry from monocular video on mobile
  {GPU}s.
\newblock In {\em Third Workshop on Computer Vision for AR/VR, Long Beach, CA},
  2019.

\bibitem{crema}
Michael~K Keutmann, Samantha~L Moore, Adam Savitt, and Ruben~C Gur.
\newblock Generating an item pool for translational social cognition research:
  methodology and initial validation.
\newblock {\em Behavior research methods}, 47(1):228--234, 2015.

\bibitem{kim2018DeepVideo}
H. Kim, P. Garrido, A. Tewari, W. Xu, J. Thies, N. Nie{\ss}ner, P. P{\'e}rez,
  C. Richardt, M. Zollh{\"o}fer, and C. Theobalt.
\newblock {Deep Video Portraits}.
\newblock {\em ACM TOG}, 2018.

\bibitem{kr2019towards}
Prajwal KR, Rudrabha Mukhopadhyay, Jerin Philip, Abhishek Jha, Vinay
  Namboodiri, and CV Jawahar.
\newblock Towards automatic face-to-face translation.
\newblock In {\em ACM Multimedia}, pages 1428--1436, 2019.

\bibitem{EAheadstab}
Mathieu Lamarre, J.P. Lewis, and Etienne Danvoye.
\newblock Face stabilization by mode pursuit for avatar construction.
\newblock In {\em Image and Vision Computing}, pages 1--6. {IEEE}, 2018.

\bibitem{blendshapesurvey14}
J.P. Lewis, Ken Anjyo, Taehyun Rhee, Mengjie Zhang, Fr{\'{e}}d{\'{e}}ric~H.
  Pighin, and Zhigang Deng.
\newblock Practice and theory of blendshape facial models.
\newblock In Sylvain Lefebvre and Michela Spagnuolo, editors, {\em Eurographics
  - State of the Art Reports}. Eurographics Association, 2014.

\bibitem{flame}
Tianye Li, Timo Bolkart, Michael~J Black, Hao Li, and Javier Romero.
\newblock Learning a model of facial shape and expression from {4D} scans.
\newblock {\em ACM TOG}, 36(6):194, 2017.

\bibitem{mathieu2015deep}
Michael Mathieu, Camille Couprie, and Yann LeCun.
\newblock Deep multi-scale video prediction beyond mean square error.
\newblock {\em ICLR}, 2016.

\bibitem{wacv}
Gaurav Mittal and Baoyuan Wang.
\newblock Animating face using disentangled audio representations.
\newblock In {\em WACV}, 2019.

\bibitem{cpbd}
Niranjan~D Narvekar and Lina~J Karam.
\newblock A no-reference perceptual image sharpness metric based on a
  cumulative probability of blur detection.
\newblock In {\em International Workshop on Quality of Multimedia Experience},
  pages 87--91. IEEE, 2009.

\bibitem{wav2lip}
KR Prajwal, Rudrabha Mukhopadhyay, Vinay~P Namboodiri, and CV Jawahar.
\newblock A lip sync expert is all you need for speech to lip generation in the
  wild.
\newblock In {\em Proceedings of the 28th ACM International Conference on
  Multimedia}, pages 484--492, 2020.

\bibitem{linsen2020ebt}
Linsen Song, Wayne Wu, Chen Qian, Chen Qian, and Chen~Change Loy.
\newblock Everybody’s talkin’: Let me talk as you want.
\newblock {\em arXiv preprint}, arXiv:, 2020.

\bibitem{obama}
Supasorn Suwajanakorn, Steven~M Seitz, and Ira Kemelmacher-Shlizerman.
\newblock Synthesizing {O}bama: learning lip sync from audio.
\newblock {\em ACM (TOG)}, 36(4):95, 2017.

\bibitem{thies2020neural}
Justus Thies, Mohamed Elgharib, Ayush Tewari, Christian Theobalt, and Matthias
  Nie{\ss}ner.
\newblock Neural voice puppetry: Audio-driven facial reenactment.
\newblock In {\em ECCV}, pages 716--731. Springer, 2020.

\bibitem{face2face}
Justus Thies, Michael Zollhofer, Marc Stamminger, Christian Theobalt, and
  Matthias Nie{\ss}ner.
\newblock Face2face: Real-time face capture and reenactment of rgb videos.
\newblock In {\em CVPR}, pages 2387--2395, 2016.

\bibitem{Umeyama}
Shinji Umeyama.
\newblock Least-squares estimation of transformation parameters between two
  point patterns.
\newblock {\em IEEE TPAMI}, 13(4):376--380, 1991.

\bibitem{vougioukas2019end}
Konstantinos Vougioukas, Stavros Petridis, and Maja Pantic.
\newblock End-to-end speech-driven realistic facial animation with temporal
  gans.
\newblock In {\em CVPR Workshops}, pages 37--40, 2019.

\bibitem{ijcv19}
Konstantinos Vougioukas, Stavros Petridis, and Maja Pantic.
\newblock Realistic speech-driven facial animation with {GAN}s.
\newblock {\em IJCV}, pages 1--16, 2019.

\bibitem{wang2011text}
Lijuan Wang, Wei Han, Frank~K Soong, and Qiang Huo.
\newblock Text driven {3D} photo-realistic talking head.
\newblock In {\em Twelfth Annual Conference of the International Speech
  Communication Association}, 2011.

\bibitem{wikiSTFT}
Wikipedia.
\newblock Short-time {F}ourier transform.
\newblock \url{https://en.wikipedia.org/wiki/Short-time_Fourier_transform},
  2019.

\bibitem{x2face}
Olivia Wiles, A Sophia~Koepke, and Andrew Zisserman.
\newblock X2face: A network for controlling face generation using images,
  audio, and pose codes.
\newblock In {\em ECCV}, pages 670--686, 2018.

\bibitem{williams:recurrent}
R.~J. {Williams} and D. {Zipser}.
\newblock A learning algorithm for continually running fully recurrent neural
  networks.
\newblock {\em Neural Computation}, 1(2):270--280, 1989.

\bibitem{yang2014efficient}
Qingxiong Yang, Jinhui Tang, and Narendra Ahuja.
\newblock Efficient and robust specular highlight removal.
\newblock {\em IEEE TPAMI}, 37(6):1304--1311, 2014.

\bibitem{aaai}
Hang Zhou, Yu Liu, Ziwei Liu, Ping Luo, and Xiaogang Wang.
\newblock Talking face generation by adversarially disentangled audio-visual
  representation.
\newblock In {\em AAAI}, volume~33, pages 9299--9306, 2019.

\bibitem{zhou2018visemenet}
Yang Zhou, Zhan Xu, Chris Landreth, Evangelos Kalogerakis, Subhransu Maji, and
  Karan Singh.
\newblock Visemenet: Audio-driven animator-centric speech animation.
\newblock {\em ACM TOG}, 37(4):161, 2018.

\end{thebibliography}

\pagebreak
\section*{Appendix}
\appendix
\section{User Study Analysis}
\label{sec_user_study_analysis}

\newcommand{\TBEref}{\cite{fried2019text}}
\newcommand{\NVPref}{\cite{thies2020neural}}

We conducted a user study to quantitatively compare our lip-sync and perceptual quality against the state-of-the-art audio-driven frameworks of Wav2Lip \cite{wav2lip}, 
NVP \NVPref, IJCV'19 \cite{ijcv19} and TBE \TBEref.  
In the study, N=35 raters were each shown a total 29 sample clips consisting of synthetic and real videos. For competing methods, we used their released videos (NVP, TBE) or generated results with their pre-trained models (IJCV'19, Wav2Lip). 
The raters were primarily drawn from a pool of subjects without research expertise,
supplemented with a minority (N=14) who were researchers. 
The subgroup of researchers included some having familiarity with computer vision topics but none were expert on speech-driven animation.
The raters were asked three questions: 
\textbf{Q1:} \textit{Is the video real or fake?}
\textbf{Q2:} \textit{Rate the quality of the lip-sync, i.e.~how well does the motion of the lips match the audio, on a 3-point (discrete) scale (poor, acceptable, great).}
\textbf{Q3:} \textit{Rate the picture quality, e.g.~naturalness, resolution, and consistency of the video, on a discrete 5-point scale from 1-5 (poor-great)}.

\begin{figure*}[!ht] %----------------
\centering
\begin{subfigure}{0.45\linewidth}
\includegraphics[width=\textwidth]{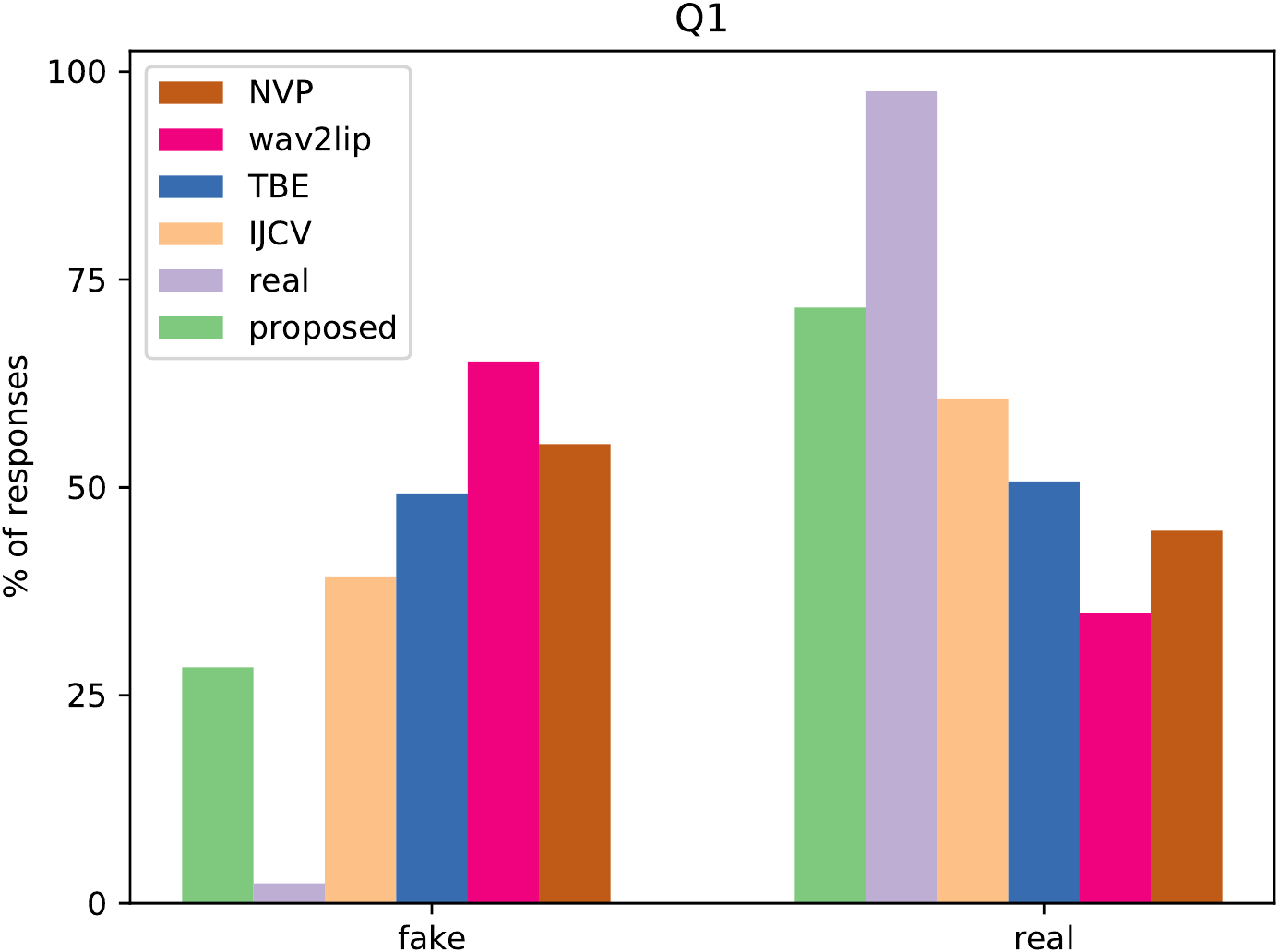}
\end{subfigure}
\hfill
\begin{subfigure}{0.45\linewidth}
\includegraphics[width=\textwidth]{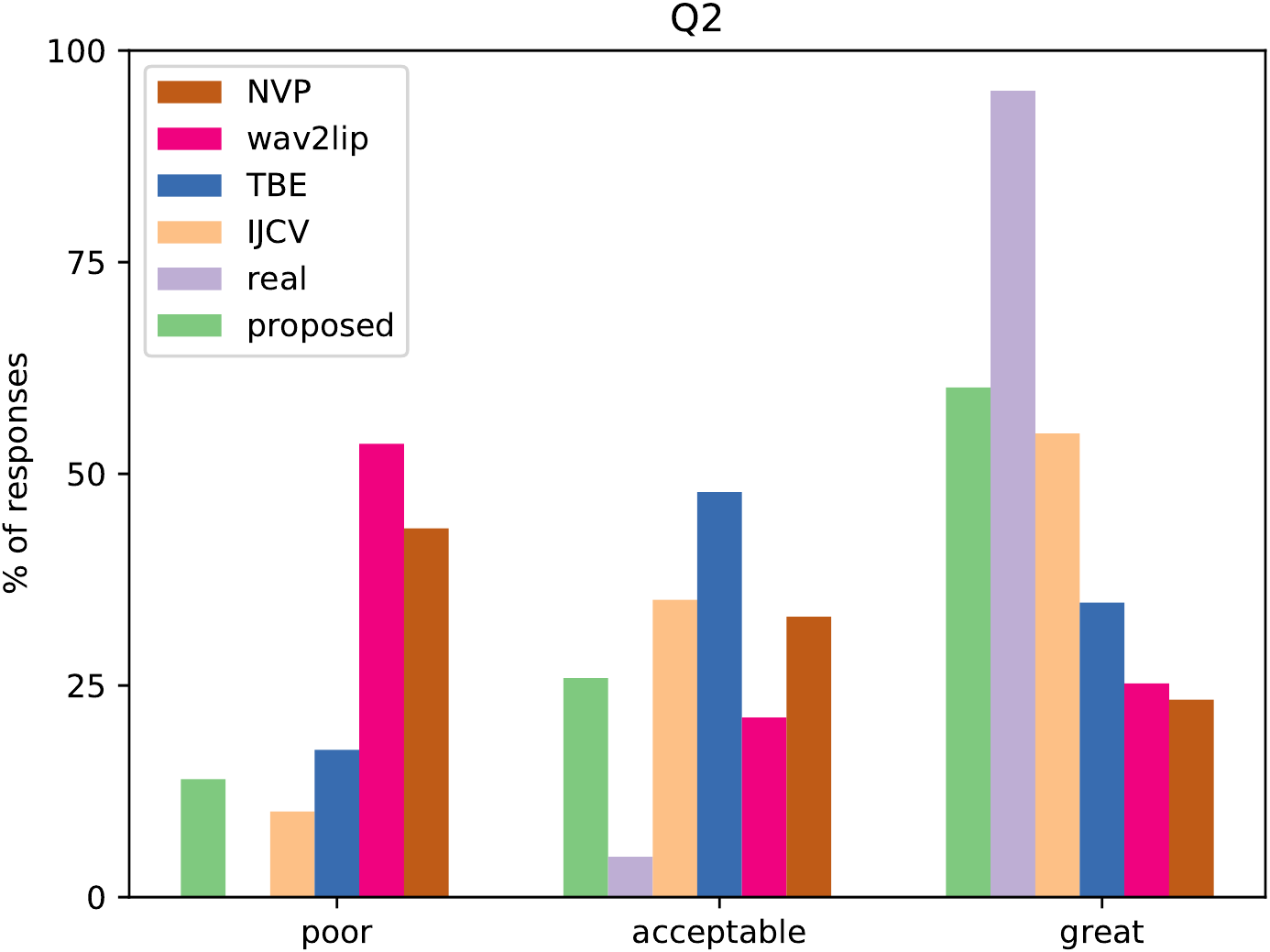}
\end{subfigure}\\

\begin{subfigure}{0.75\linewidth}
\includegraphics[width=\textwidth]{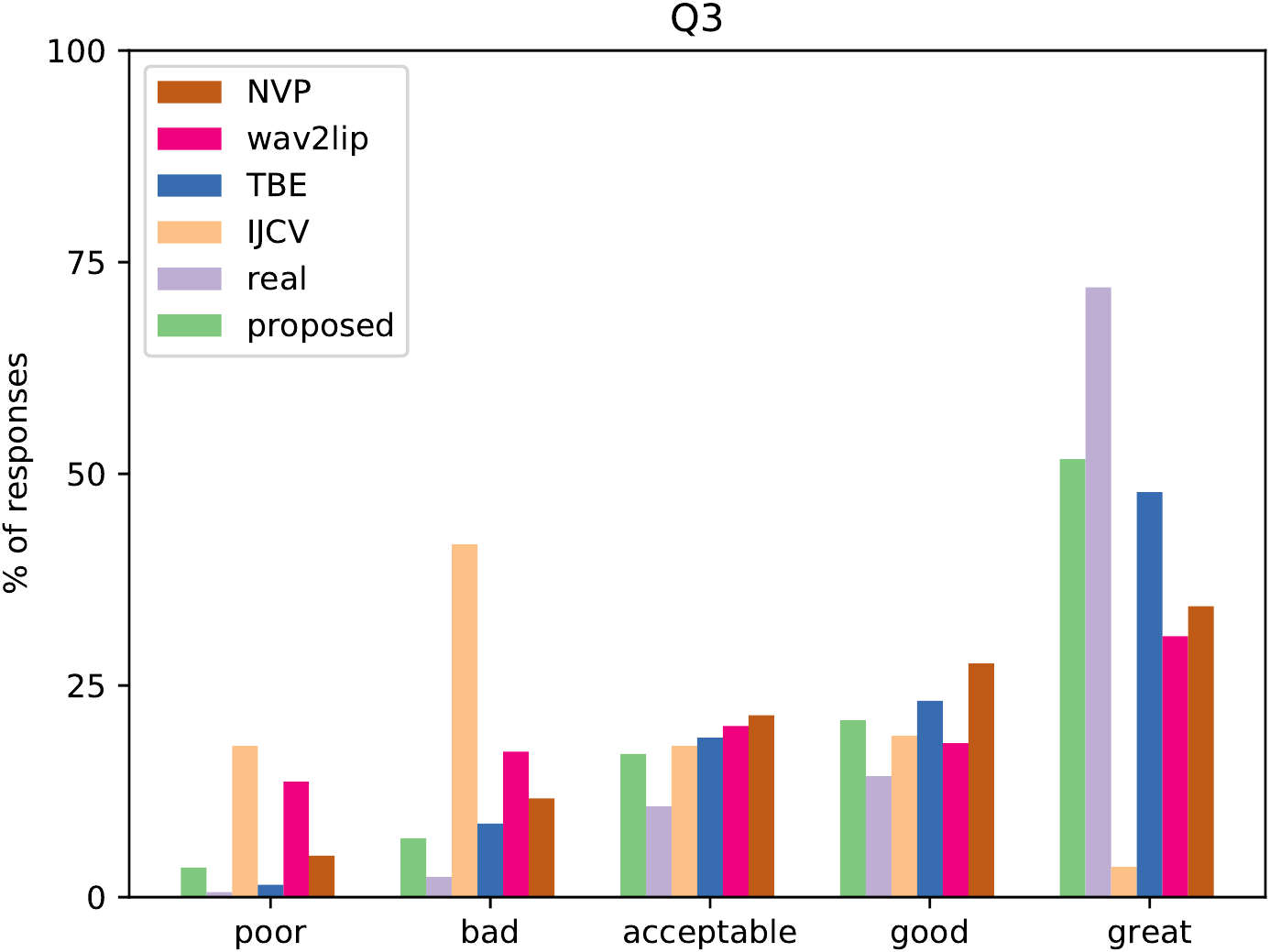}
\end{subfigure}

\vspace{1em}
\caption{
Raw user study results. \textbf{Q1}: Percentage of real/fake ratings for each of the six video categories. Viewers believe our synthetic videos (proposed) are real roughly two-thirds of the time.
\textbf{Q2}: Ratings of lip-sync quality for the six video categories, expressed as percentages. Our synthetic videos (proposed) are perceived as at least comparable to those of IJCV while being clearly superior to other competing methods in lip-sync quality.
\textbf{Q3}: Ratings of picture quality for the six video categories, expressed as percentages. 
Here our method greatly outperforms IJCV (and NVP and wav2lip) while being at least comparable to TBE.
It can be seen that our method is the best performer across the three questions.
}
\label{fig:user_study_barplots}
\end{figure*}

Figure~\ref{fig:user_study_barplots} shows, for each question, the percentage of raters who selected each rating.
 We report the Mean Opinion Scores (MOS) of the questions in Table~\ref{tab:ratings}.
As is evident, among the competing methods, our method receives the most favorable user ratings.

We performed a statistical analysis to confirm the significance of these ratings.
For Q1 (only) we excluded the text-to-speech results from consideration, 
since it was straightforward to judge the videos as ``fake'' due to the computer-generated speech. However, questions Q2 and Q3 are still relevant in the text-to-speech case, since it is possible to rate the quality of lip-sync and overall image naturalness even when the voice is clearly synthetic.  

The statistical analysis confirms that the differences in real-fake ratings on Q1 are significant 
%(using a nonparametric test, Kruskal-Wallis one-way ANOVA p $<$ 1e-04) 
(Kruskal-Wallis test, $\chi^2 = 158$, p$<$1e-04),
and our method outperforms the other methods after adjusting for multiple comparisons
(Tukey's Honest Significant Differences (HSD) IJCV~p~adj.$=.003$, Wav2Lip p~adj.$<$1e-04, NVP p~adj.$<$1e-04).
For Q2 the differences in ratings are significant (Kruskal-Wallis $\chi^2=279$, p$<$1e-04), 
and our method outperforms most competing methods 
with statistical significance after adjusting for the multiple tests 
(Tukey's HSD~ TBE p~adj.$=.035$, Wav2Lip p~adj.$<$1e-04, NVP p~adj.$<$1e-04),
however the difference versus IJCV'19 is not significant. 
For Q3 (Kruskal-Wallis $\chi^2=248$, p$<$1e-04) our method outperforms most competing methods 
with statistical significance after adjustment for multiple comparison
(HSD~ IJCV p~adj.$<$1e-04, Wav2lip p~adj.$<$1e-04, NVP p~adj.$=0.04$
however the comparison with TBE is not significant.

\begin{figure}[h]
\includegraphics[width=\linewidth]{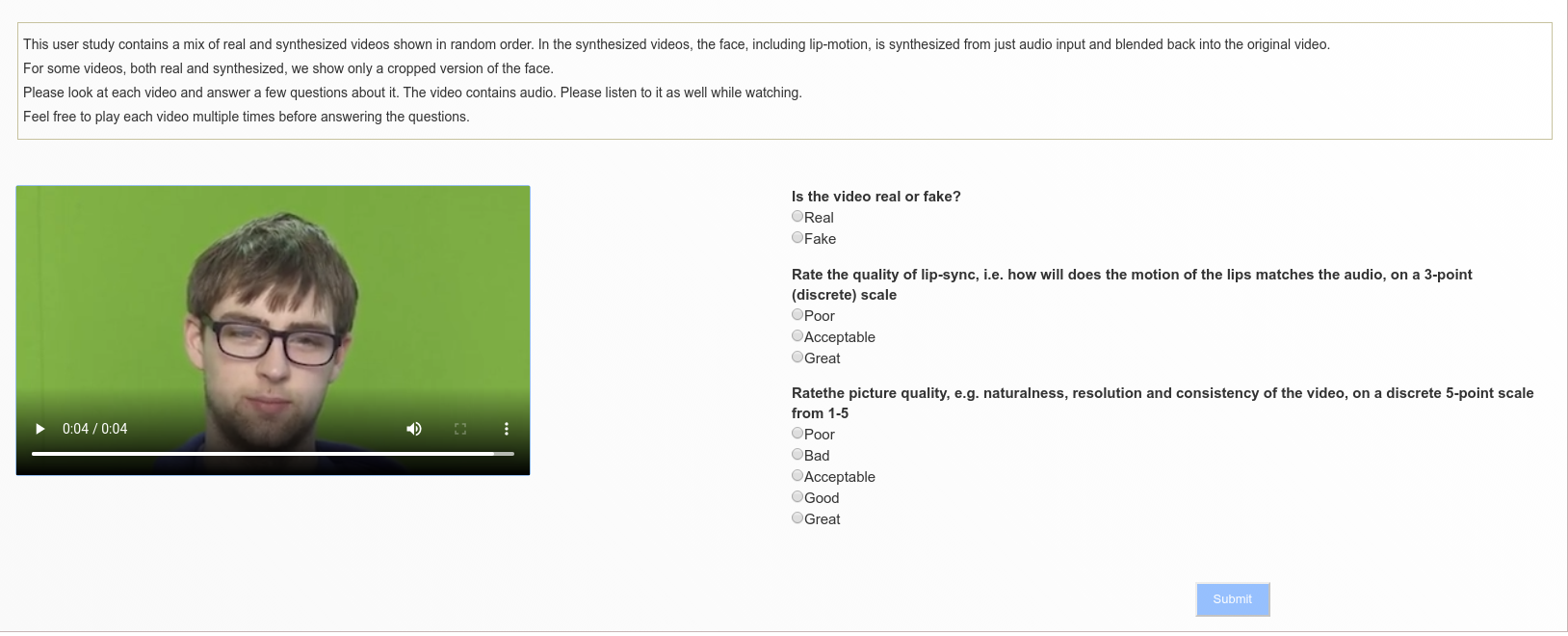}
\vspace{0.1em}
\caption{Screen shot from our user study.}
\label{fig:userstudy}
\end{figure}

 %----------------

\begin{table}  % [h!]
\centering
\scriptsize
\begin{tabular}{|lcccc|} \hline \hline
\multicolumn{1}{|c}{Method} &
\multicolumn{1}{c}{\begin{tabular}[c]{@{}c@{}}Is Real ?\\ (\% Yes) \end{tabular}} &
\multicolumn{1}{c}{\begin{tabular}[c]{@{}c@{}}[no TTS]\\ Is Real ?\\ (\% Yes) \end{tabular}} &
  \multicolumn{1}{c}{\begin{tabular}[c]{@{}c@{}}Lip-Sync\\ (1-3) )\end{tabular}} &
  \multicolumn{1}{c|}{\begin{tabular}[c]{@{}c@{}}Visual Quality\\ (1-5)\end{tabular}} \\ \hline
Real    & 97.6  & 97.6  & 2.95$\pm$0.03 & 4.55$\pm$0.13 \\ \hline
IJCV'19 & 60.7  & 60.7  & 2.45$\pm$0.11 & 2.49$\pm$0.17  \\
Wav2Lip  & 34.4 & 37.6  & 1.72$\pm$0.12 & 3.35$\pm$0.20 \\
NVP     & 44.4   & 50.0  & 1.80$\pm$0.13 & 3.75$\pm$0.19   \\
TBE    & 50.7   & n/a  & 2.17$\pm$0.17 & 4.07$\pm$0.26   \\
\textbf{Proposed} & \textbf{71.6}  & \textbf{77.25}  & \textbf{2.46$\pm$0.10} & \textbf{4.10$\pm$0.15}  \\ \hline
\end{tabular}
\vspace{1em}
\caption{User study analysis. Column 1: percentage of ``real'' ratings by category. Column 2: percentage of ``real'' ratings with text-to-speech driven results removed. 
Column 3: mean opinion score of lip-sync quality.
Column 4: mean opinion score of picture quality.
}
\label{tab:ratings}
\end{table}

These significance results for Q2 and Q3 (and in particular the lack of significance for IJCV and TBE respectively) are plausible given cursory examination of the videos.
The results of IJCV'19 show qood quality lip-sync but the overall image quality is limited,
thus explaining its good performance on Q2 but poor performance on Q3. TBE operates in part by re-mixing input video frames so it results in high picture quality by definition (Q3), but its lip-sync quality is poorer than our method and that of IJCV'19. 

\section{Comparison Notes on Text Based Editing~\cite{fried2019text}}
According to the user study, among the 3D model based methods, Text-based-Editing (TBE) is the second-best method (following our method). However, our framework has some distinct training and inference time advantages over TBE:
\begin{itemize}
    \item TBE was trained on a training corpus of more than 1 hour of video recording. Our model was trained on $\sim$7 minutes of data in this case.
    \item TBE assumes an accurate text transcript and uses phoneme based alignment tools to align the text with audio. In contrast, our model only requires a speech signal as input.
    \item The average training time of TBE is 42 hours. Our typical training time is somewhere in between 3-5 hours.
    \item The inference speed of TBE is significantly slower. This is mainly attributed to the costly viseme search ($\sim$5 minutes for 3 words). Our method executes within a few tens of milliseconds. 
    \item TBE also relies on neural rendering for learning to generate facial texture based on the illumination in the training sequence. It is thus not apt for operating under new ambient lighting without further retraining. Our framework is capable of seamlessly adapting to novel lighting conditions during inference. 
\end{itemize}
\section{Limited comparison with Suwajanakorn \textit{et al.}~\cite{obama}}
The work by Suwajanakorn \textit{et al.}~also involves training a personalized talking face model. However, \cite{obama} only synthesized results for a single person (former U.S.~President Barack Obama), using hours of training video. While our model is perfectly capable of similar synthesis, we consciously refrain from training on living political personalities due to ethical considerations. Hence, we are unable to directly compare our work with that of Suwajanakorn \textit{et al.}. Nevertheless, our framework offers the following advantages:
\begin{itemize}
    \item Suwajanakorn \textit{et al.}~trained on 14 hours of weekly President addresses recorded between 2009-2016. In contrast, our framework just requires $\sim$5 minutes of training video.
    \item Suwajanakorn \textit{et al.}~demonstrate their outputs only under the specific studio lighting setup of the President's office. Their texture generation 
    network is not designed with the goal of handling diverse ambient lighting. In contrast, our network disentangles and normalizes the effects of illumination, thereby enabling inference under diverse lighting conditions.
    \item Typical training data pre-processing time of Suwajanakorn \textit{et al.} is around 2 weeks on 10 cluster nodes of Intel Xeon E5530. In contrast, our combined pre-processing and training takes only about 3-5 hours on a single system equipped with a NVIDIA P1000 GPU. 
\end{itemize} 
\section{Discussion on LSE metrics}
We have used the official code release \footnote{\href{https://github.com/Rudrabha/Wav2Lip}{https://github.com/Rudrabha/Wav2Lip}} by the authors of Wav2Lip for evaluating the automated LSE metrics, LSE-D (lower is better) and LSE-C (higher is better). We faced two issues while using this metric:\\
\textbf{(a:)~} Even though user study suggest that the lip-sync quality of Wav2Lip is usually inferior to our model, the LSE metrics are always better for Wav2Lip. We observed this pattern over a range of different videos. 
LSE metrics are computed from paired audio-visual representation coming from a SyncNet~\cite{syncnet} network. Wav2Lip also leverages a SyncNet architecture and audio-visual representations as a lip-sync loss during training. 
We hypothesize that since a similar architecture is used both as a training loss and for scoring,
 Wav2Lip may be biased to do particularly well on this metric.
Thus, we refrained from reporting LSE metrics for Wav2Lip. However, for other methods, the metric yields numbers consistent with human evaluations of lip-sync.\\
\textbf{(b:)~} The code sometimes fails to detect faces even under normal illumination and thereby does not give LSE metrics. So, we could not report LSE metrics on all videos. 
 \section{Leaving out Wav2Lip for Self-reenactment Comparisons}
While comparing methods for self-reenactment tasks (\FIGREF{fig_combined_result}), we do not include Wav2Lip among the competing methods. Along with the current audio, Wav2Lip also feeds in the sequence of target frames with the lip region unmasked.
In a self-reenactment setting, the input target frames are same as the final expected output from the network. Hence, Wav2Lip would have an unfair advantage over our method, since our framework is entirely audio driven and only utilizes masked target frames (around the lip region) for pasting back the synthesized output.
Therefore, we do not compare Wav2Lip for self-reenactment results. A similar advantage is also available to LipGAN~\cite{kr2019towards} which is a precursor to the framework of Wav2Lip.

\begin{figure}
 \vspace{1em}
    \centering
    \includegraphics[scale=0.22]{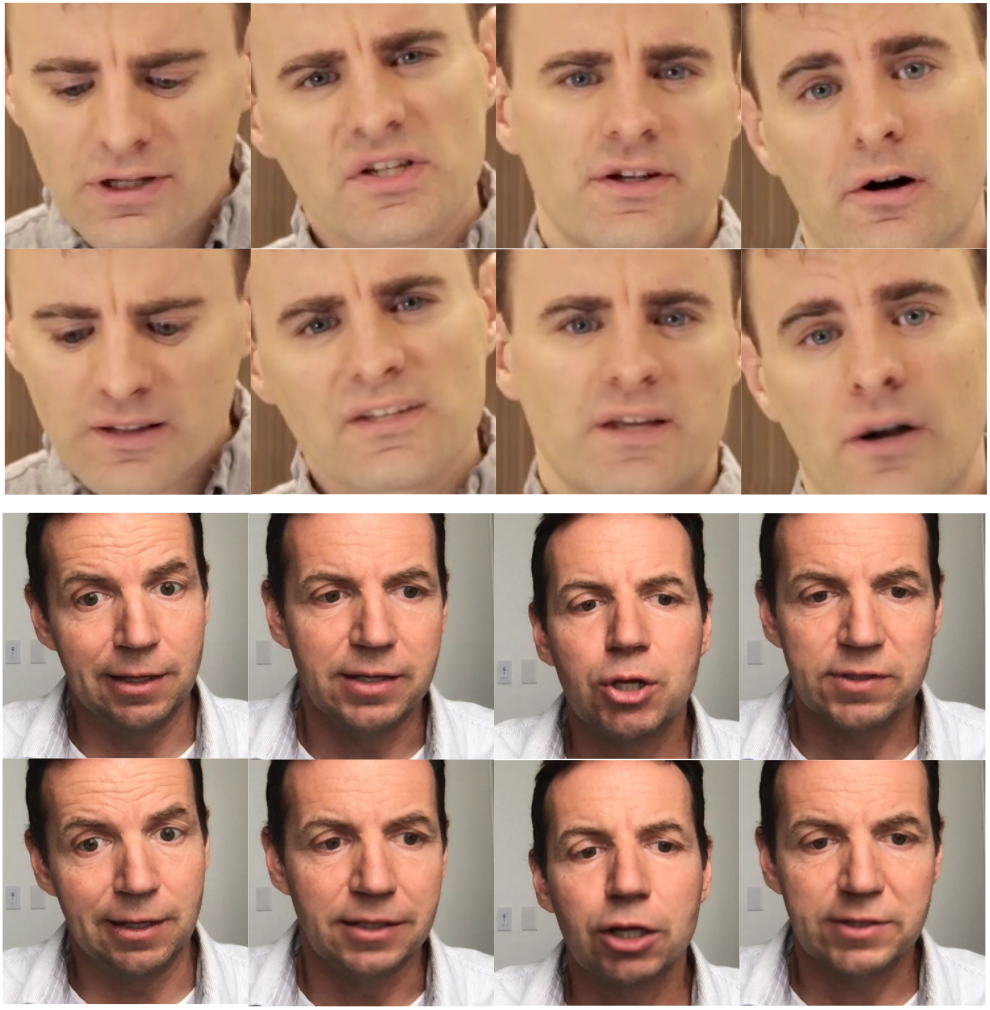}
    \caption{ Comparing our result against ground-truth. For each subject, top row is the original sequence of frames, while the bottom row is the resynthesized sequence.}
    \label{fig_sequence}
\end{figure} % moved figure earlier

\section{Selecting Code Length for Audio and Previous Atlas}
In this section we report studies to determine the code length for encoding the current time step's spectrogram and previous time step's predicted atlas. Since we wish to automatically determine acceptable settings of these parameters, the study was conducted for a self-reenactment task in which we have access to ground truth facial information.

\par For this study, we curated a custom dataset of subjects selected from YouTube instructional videos, webcam recordings, and studio conversations. The custom dataset had around 10,000 audio-synchronized frames.
Sample frames and corresponding reconstructions from two such subjects are shown in~\FIGREF{fig_sequence}.

\subsection{Selection of Previous Atlas Code Length, $N_a$}
We conducted an ablation study to determine the length $N_a$ of the latent code $L^A_{t-1}$. The code length governs the contribution of the previous visual state to the current frame. With increasing code length, the model starts to incorrectly neglect the current audio input, instead basing its output mostly on the previous state. In Table~\ref{tab_ablation_code_length} we report the average metrics over different $N_a$. Note that $N_a$ = 0 signifies a model trained without auto-regression. Both SSIM and LMD improve when using auto-regression initially but deteriorate as we increase $N_a$, with $N_a=2$ giving the best results.

\begin{table}[ht]
\centering
\vspace{1em}
\begin{tabularx}{\linewidth}{|X|XXXX|} \hline \hline 
Metric & \multicolumn{4}{|c|}{$\leftarrow$ AR Code Length: $N_a$ $\rightarrow$}  \\\hline
 & 0 & 2 & 8 & 16 \\ \hline
SSIM & 0.92 & \textbf{0.93} & 0.901 & 0.889 \\
LMD & 2.00 & \textbf{1.88} & 2.91 & 2.16 \\ \hline
\end{tabularx}
\vspace{0.7em}
\caption{Parameter sweep for selecting latent code lengths, $N_a$, for encoding previous time step atlas based on LMD (lower is better) and SSIM (higher is better) metrics. Best results are marked in bold.}
\label{tab_ablation_code_length}
\end{table}
%=======================================================
\subsection{Selection of Audio Code Length, $N_S$}
As mentioned earlier, we used a 32-dimensional vector for encoding the audio spectrogram. This number was chosen by performing a parameter sweep over $N_S \in \{8, 16, 32, 64, 128\}$ on our custom dataset. In Table~\ref{tab_ablation_audio} we compare the SSIM and LMD metrics, averaged over the subjects in the dataset. We observe that $N_S = 32$ yields the most favorable performance. Hence, we use it as our default choice when encoding the spectrogram.
%================ Tabble Audio Ablation Starts =================
\begin{table}[ht]
\centering
\begin{tabularx}{\linewidth}{|X|XXXXX|}\hline \hline 
 & \multicolumn{5}{|c|}{$\leftarrow$ Audio Code Length: $N_S \rightarrow$} \\\hline\hline
Metric & 8 & 16 & 32 & 64 & 128 \\\hline
LMD & 1.83 & 1.87 & \textbf{1.77} & 1.81 & 1.82 \\
SSIM & 0.907 & 0.914 & \textbf{0.917} & 0.917 & 0.910\\ \hline
\end{tabularx}
\vspace{0.5em}
\caption{Parameter sweep for selecting audio latent code length based on LMD and SSIM metrics. Best results are marked in bold.}
\label{tab_ablation_audio}
\end{table}
%================= Table Audio Ablation Ends=============

\section{Network Architectures}
In the main paper, we describe the architecture of the audio encoder, which computes the latent code from audio spectrograms, and the geometry decoder, which computes the 3D vertices from the audio latent code. Here, we describe the additional network components of our model.

\subsection{Texture Decoder Architecture}
\label{sec_texture_decoder}
In Table~\ref{tab_decoder} we present  details of the texture decoder. The input to the decoder is either a 32D vector (conditioned on only audio latent code), or 34D (conditioned on audio latent code + previous time step atlas latent code). This is followed by a series of convolution and upsampling layers.

%=========== Architecture of Decoder=============
\begin{table*}[!h]
\centering
\begin{tabular}{|l|lllll|}\hline \hline 
Input & Type & ~~Kernel & ~~Stride & ~~Channels & ~~Outputs \\\hline
\multicolumn{6}{|l|}{\begin{tabular}[c]{@{}l@{}}Input (Latent Vector):\\ = 32D (only audio)\\ = 34D (Audio + AutoRegressive)\end{tabular}} \\\hline
Latent Vector & FC &~~~~ - &~~~~ - &~~~~ - &~~~~ 16384 \\
\multicolumn{6}{|l|}{Reshape: 4$\times$4$\times$1024} \\
\multicolumn{6}{|l|}{2$\times$ Bilinear Upsample} \\
8$\times$8$\times$1024 & Conv &~~~~ 3$\times$3 &~~~~ 1$\times$1 &~~~~ 512 &~~~~ 8$\times$8$\times$512 \\
\multicolumn{6}{|l|}{2$\times$ Bilinear Upsample} \\
16$\times$16$\times$512 & Conv &~~~~ 3$\times$3 &~~~~ 1$\times$1 &~~~~ 256 &~~~~ 16$\times$16$\times$256 \\
\multicolumn{6}{|l|}{2$\times$ Bilinear Upsample} \\
32$\times$32$\times$256 & Conv &~~~~ 3$\times$3 &~~~~ 1$\times$1 &~~~~ 128 &~~~~ 32$\times$32$\times$128 \\
\multicolumn{6}{|l|}{2$\times$ Bilinear Upsample} \\
64$\times$64$\times$128 & Conv &~~~~ 3$\times$3 &~~~~ 1$\times$1 &~~~~ 64 &~~~~ 64$\times$64$\times$64 \\
\multicolumn{6}{|l|}{2$\times$ Bilinear Upsample} \\
128$\times$128$\times$64 & Conv &~~~~ 5$\times$5 &~~~~ 1$\times$1 &~~~~ 3 &~~~~ 128$\times$128$\times$3\\\hline
\end{tabular}
\vspace{0.6em}
\caption{Architecture of the texture decoder. Each fully connected (FC) and convolution layer is followed by a ReLU  non-linearity, while only the last convolution layer is followed by a tanh non-linearity. The length of the input latent vector depends on the mode of the experiment.}
\label{tab_decoder}
\end{table*}
%=============== Architecture of Decoder=============
%====================================================

\subsection{Auto-regressive Encoder Architecture}
\label{sec_autoregressive_encoder}
In our auto-regressive architecture, 
the previous atlas at the last time step is encoded as an additional latent vector (along with the audio encoded vector). The output is a latent vector of length $N_a = 2$. The encoder architecture for the auto-regressive model is shown in Table \ref{tab_encoder_ar}.\\

\noindent \textbf{Training by ``Teacher Forcing":} 
As stated in the main paper, during training we do not provide the actual previous predicted atlas as input to the auto-regressive model, since that would entail a recursion in the network. Instead we follow the \textit{Teacher Forcing}~\cite{williams:recurrent} paradigm of training the network with the ground truth previous atlas.

In our initial experiments, we implemented a recursive network and fed in the actual predicted previous atlas to the model during training. The reconstruction quality of that approach was worse than using ground truth atlases during training (\ie Teacher Forcing), however. Note that during inference, the \textit{predicted} previous atlas is fed to the model, because the ground truth atlas is not known at that time. Also, for predicting the first frame, we provide an \textit{`all-zeros'} image as a proxy for previous frame because there is no previous frame to start with. To handle this case, we train the model by feeding it with \textit{`all-zeros'} for the previous atlas with a probability of $20\%$. This trains the model to reconstruct the atlas both with and without the knowledge of previous time step's visual state.

%==========Architecture of AR Encoder======
\begin{table*} [!h]
\centering
\begin{tabular}{|llcccc|}\hline \hline 
Input & Type & Kernel & Stride & Channels & Outputs \\\hline
\multicolumn{6}{|l|}{\begin{tabular}[c]{@{}l@{}}Input (RGB):\\ =128$\times$128$\times$3\end{tabular}} \\\hline
Input & Conv & 5$\times$5 & 2$\times$2 & 128 & 64$\times$64$\times$128 \\
64$\times$64$\times$128 & Conv & 5$\times$5 & 2$\times$2 & 256 & 32$\times$32$\times$256 \\
32$\times$32$\times$256 & Conv & 5$\times$5 & 2$\times$2 & 512 & 16$\times$16$\times$512 \\
16$\times$16$\times$512 & Conv & 5$\times$5 & 2$\times$2 & 1024 & 8$\times$8$\times$1024 \\
8$\times$8$\times$1024 & Conv & 5$\times$5 & 2$\times$2 & 2048 & 4$\times$4$\times$2048 \\
4$\times$4$\times$2048 & FC & - & - & - & 2\\ \hline
\end{tabular}
\vspace{0.6em}
\caption{Architecture of the encoder for the previous atlas in auto-regressive mode. The encoder input is an RGB image and output is a latent vector. Each convolution layer is followed by a ReLU non-linearity. The last fully-connected (FC) layer is followed by a tanh non-linearity.}
\label{tab_encoder_ar}
\end{table*}
%============= Architecture of AR encoder==========
\section{Subject Details: GRID, CREMA-D and TCD-TIMIT}
\label{sec_subject_details}
\vspace{-0.30em}
For self-reenactment studies we performed experiments on GRID \cite{Cooke2006}, TCD TMIT \cite{tcd} and CREMA-D \cite{crema} datasets.
Following the exact setting in \cite{ijcv19}, we select the same set of 10 subjects (see Table~\ref{tab:subject_ids}) from each of the datasets).
\\

\begin{table*}%[ht]
\centering
\begin{tabularx}{.7\linewidth}{|c|X|}\hline \hline
Dataset & Test Subject ID \\\hline
GRID & 2, 4, 11, 13, 15, 18, 19, 25, 31, 33 \\
TCD-TMIT~~ & 8, 9, 15, 18, 25, 28, 33, 41, 55, 56 \\
CREMA-D & 15, 20, 21, 30, 33, 52, 62, 81, 82, 89\\ \hline
\end{tabularx}
\vspace{0.5em}
 \caption{IDs of subjects used for self-reenactment experiment.}
\label{tab:subject_ids}
\end{table*}

\balance
\vspace{-0.90em}
\section{Sharpness of Synthesized Lip Region}
In main paper, we mentioned that our method is capable of generating high quality lip-sync, and we objectively established this with the commonly used CPBD metric. The metric was evaluated on the entire face. While it is true that the sharpness of the final composite full face is a primary factor of photo-realism, we also acknowledge that our method benefits from copying the texture from upper part of face from target frames.
\par Here we focus on determining the sharpness of only the lower half of the face (below the nostrils). In Table \ref{tab_lip_region}, we compare against Wav2Lip, LipGAN, NVP and TBE on the user study videos. Even on the lower mouth region, our method attains better CPBD scores across all competing methods.

\begin{table*}%[!t]
\centering
\begin{tabular}{|cc|cc|cc|cc|}\hline  \hline
LipGAN & Proposed & Wav2Lip & Proposed & TBE  & Proposed & NVP  & Proposed \\\hline
0.07   & 0.14     & 0.06    & 0.13     & 0.12 & 0.18     & 0.10 & 0.18   \\\hline 
\end{tabular}
\vspace{0.6em}
\caption{Comparing CPBD metrics (on lower half of faces) of competing methods against our proposed method. For each method, we select common pairs of videos for the competing and our method from the pool of user study videos.}
\label{tab_lip_region}
\end{table*}
%\FloatBarrier

\section{Applications}
\label{sec_supple_applications}

\begin{figure*}[!t]
%\vspace{-0.6em}  %jp/space
    \centering
    \includegraphics[width=\textwidth]{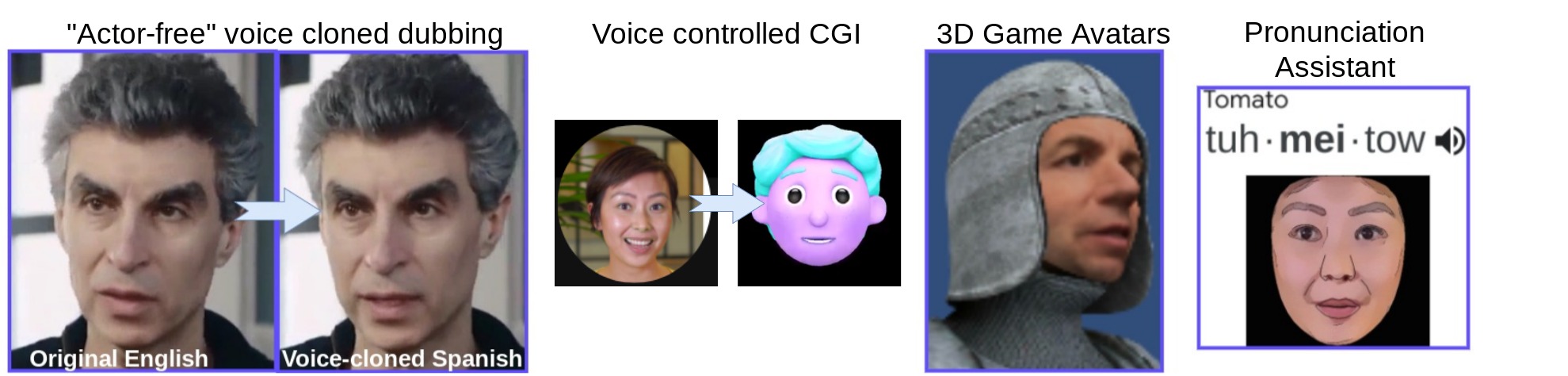}
    %\vspace{0.1cm}
    \caption{Sample applications enabled by our 3D talking face generation pipeline.}
    \label{fig_applications}
\end{figure*}

Our approach of generating textured 3D geometry enables us to address a broader variety of applications than purely image-based or 3D-only techniques, as discussed here. Sample screenshots from some of these applications are shown in~\FIGREF{fig_applications}.

\paragraph{3D Talking Avatars:}
3D avatars can make multiplayer online games and Virtual Reality (VR) environments more social and engaging. They may also be employed for audio/video chat applications and virtual visual assistants. While such avatars can be driven by a video feed from a web-cam or head-mounted camera, the ability to generate a 3D talking face from just audio obviates the need for any auxiliary camera device, and also helps preserve privacy, while reducing bandwidth requirements at the same time. Our technique supports generating both 3D textured faces as well as CGI avatars for these applications. 

\textbf{Video creation and editing:} Our approach can be used for editing videos,~\eg to insert new content in an online course, or to correct an error without the cumbersome and sometimes impossible procedure of re-shooting the whole video under original conditions. Instead, a new audio transcript may be recorded for the edited portion, followed by applying our synthesis technique to modify the corresponding video segment. Such a speech-to-video or text-to-video system may be useful in multiple domains such as education, advertising and entertainment. We can also generate cartoon renderings for these videos, which may be preferred in some applications,~\eg sketch videos, stylized animations, or assistive technologies such as pronunciation visualization.\\

\textbf{Video translation and dubbing:}
Even though we train our models on videos in a single language, they are surprisingly robust to both different languages as well as text-to-speech (TTS) audio at inference time. Using available transcripts or a speech recognition system to obtain captions, and subsequently a text-to-speech system to generate audio, we can automatically translate and lip-sync existing videos into different languages. In conjunction with appropriate video re-timing and voice-cloning~\cite{hsu2019_voicecloning}, the resulting videos look fairly convincing. We have employed our approach for translating and dubbing videos from English to Spanish or Mandarin, and vice-versa. Notably, in contrast to narrator-driven techniques~\cite{kim2018DeepVideo,face2face}, our approach for video dubbing does not require a human actor in the loop, and is therefore more scalable across languages.

\end{document}

%%%%%%%%% TITLE
\title{3D Photorealistic Talking Faces from Audio \\ (Supplementary)}

\author{First Author\\
Institution1\\
Institution1 address\\
{\tt\small firstauthor@i1.org}
\and
Second Author\\
Institution2\\
First line of institution2 address\\
{\tt\small secondauthor@i2.org}
}

\maketitle
\section{Network Architectures}
\subsection{Texture Prediction Network}
Table \ref{tab_audio_encoder} specifies  the architecture of the input spectrum encoder network.
%In Table \ref{tab_audio_encoder} we specify the network details for the encoder network to encode the input spectrogram. 
In Table \ref{tab_encoder_lighting} we list the network details for encoding the lighting atlas. The encoder network for encoding the previous atlas in auto-regressive (AR) mode is similar to that of the lighting encoder except that the last fully connected layer has two nodes for the AR encoder instead of four for the lighting encoder.
In Table \ref{tab_decoder} we present the network details for the decoder portion of our network to reconstruct the textured mouth region based on the input latent vector. The input latent vector for the decoder can originate from just the input spectrogram, or a combination of the encoded previous atlas (AR mode) or the encoded lighting atlas (lighting mode).

\par All losses were minimized with mini batch stochastic gradient descent optimization with the Adam optimizer. The batch size was 50 and, in general, we trained for up to 10$^5$ mini-batch iterations. The initial learning rate was 10$^{-4}$ with a decay of 20\% every 30000 steps. $\alpha_1$, which is the multiplicative coefficient of moment loss, is set to 10$^{-2}$ and $\alpha_2$, which is the multiplicative coefficient of auto-encoder loss, is set to 10$^{-3}$.
%=====================
\subsection{Vertex Prediction Network}
The audio encoder for vertex prediction has the exact same architecture as that used for texture prediction. So, given an input spectrogram, we get a 256D latent vector. The decoder section for vertex prediction \jp{is described} in the main paper.
%has already being discussed
%============================================
\section{Experiments}
\subsection{Ablation of audio latent vector}
We performed a \jp{hyper-parameter} search for determining the length, $L$, of the latent vector to encode the audio spectrograms. We inspected the Landmark Distance (LMD) metric to select the best latent code length. The LMD metrics averaged over Speakers A and B (same identities as referred to in main paper) are reported in Table \ref{tab_lmd_with_audio}. $L$=256 achieves the \jp{least} error. 
%===================================
\subsection{Additional experiment on GRID dataset}
In Table \ref{tab_grid} we report the LMD and SSIM on subjects s1, s2, s3 and s29 on the GRID dataset. We used 50\% of the sentences from each subject for training, and tested on the remaining 50\%. It is not possible to do an apples-to-apples comparison with recent talking-head frameworks of \cite{chen2018lip, x2face, chung2017you, chen2019hierarchical}. However, on average, our metrics are much superior compared to the best performing frameworks which report \cite{wacv} LMD greater than 1.1 and SSIM less than 0.8.
%===================== abaltion code length
\begin{table}[]
\centering
\begin{tabular}{llllll}\hline\hline
L=32 & L=64 & L=128 & L=256 & L=512 & L=1024 \\\hline
3.023 & 2.783 & 2.598 & 1.810 & 2.765 & 2.987\\\hline
\end{tabular}
\caption{Ablation over different lengths of latent code for the audio spectrogram. We report the average landmark detection error for each setting of $L$. Lower LMD is better. Our choice of 256D latent vector for audio gives the lowest LMD.}
\label{tab_lmd_with_audio}
\end{table}
%===========ablation code length==============
%========================= Grid compare starts========
\begin{table}[]
\centering
\begin{tabular}{lll}\hline\hline
Subject & LMD & SSIM \\\hline
s1 & 0.67 & 0.95\\
s2 & 0.68 & 0.96\\
s3 & 0.88 & 0.94\\
s29 & 0.79 & 0.93 \\\hline
\end{tabular}
\caption{Performance of our model on the GRID dataset for various subjects.}
\label{tab_grid}
\end{table}
%===========grid compare ends============

%=============== Architecture of Audio Encoder================
\begin{table*}[!t]
\centering
\begin{tabular}{llllll}\hline\hline
Input & Type & Kernel & Stride & Channels & Outputs \\\hline
256$\times$96$\times$2 & Conv & 3x1 & 2x1 &  72& 128$\times$96$\times$72 \\
 128$\times$96$\times$72&  Conv& 3$\times1$ &  2$\times$1 & 72 & 64$\times$96X72  \\
 64$\times$96$\times$72& Conv & 3$\times$1 & 2$\times$1 & 72 & 32$\times$96$\times$72 \\
 32$\times$96$\times$72& Conv & 3$\times$1 & 2$\times$1 &94  & 16$\times$96$\times$94 \\
 16$\times$96$\times$94& Conv & 3$\times$1 & 2$\times$1 & 121 & 8$\times$96$\times$121 \\
 8$\times$96$\times$121  &Conv  &3$\times$1  &2$\times$1  &158  &4$\times$96$\times$158  \\
 4$\times$96$\times$158&Conv  &3$\times$1  &2$\times$1  &205  &2$\times$96$\times$205  \\
 2$\times$96$\times$205&Conv  &3$\times$1  &2$\times$1  &256  &1$\times$96$\times$256 \\
 1$\times$96$\times$256&Conv&1$\times$3&1$\times$2&256&1$\times$48$\times$256\\
 1$\times$48$\times$256 & Conv & 1$\times$3 & 1$\times$2 & 256 & 1$\times$24$\times$256\\
 1$\times$24$\times$256 & Conv & 1$\times$3 & 1$\times$2 & 256 & 1$\times$12$\times$256\\
 1$\times$12$\times$256 & Conv & 1$\times$3 & 1$\times$2 & 256 & 1$\times$6$\times$256\\
 1$\times$6$\times$256 & Conv & 1$\times$3 & 1$\times$2 & 256 & 1$\times$3$\times$256\\
 1$\times$3$\times$256 & Conv & 1$\times$3 & 1$\times$1 & 256 & 1$\times$1$\times$256\\
 Final Output: Reshape to 256$\times$1 &&&&\\\hline
\end{tabular}
\caption{Architecture of the audio encoder. Each convolution layer is followed by a LeakyReLU \jp{non-linearity} and dropout with drop probability 0.1.}
\label{tab_audio_encoder}
\end{table*}
%================Architecture of Audio Encoder

%=========== Architecture of Decoder============
\begin{table*}[]
\centering
\begin{tabular}{l|lllll}\hline\hline
Input & Type & Kernel & Stride & Channels & Outputs \\\hline
\multicolumn{6}{l}{\begin{tabular}[c]{@{}l@{}}Input (Latent Vector):\\ = 256D (only audio)\\ = 258D (Audio + AutoRegressive)\\ = 262D (Audio + AutoRegressive + Lighting)\end{tabular}} \\\hline
Latent Vector & FC & - & - & - & 16384 \\
\multicolumn{6}{l}{Reshape: 4$\times$4$\times$1024} \\
\multicolumn{6}{l}{2$\times$ Bilinear Upsample} \\
8$\times$8$\times$1024 & Conv & 3$\times$3 & 1$\times$1 & 512 & 8$\times$8$\times$512 \\
\multicolumn{6}{l}{2$\times$ Bilinear Upsample} \\
16$\times$16$\times$512 & Conv & 3$\times$3 & 1$\times$1 & 256 & 16$\times$16$\times$256 \\
\multicolumn{6}{l}{2$\times$ Bilinear Upsample} \\
32$\times$32$\times$256 & Conv & 3$\times$3 & 1$\times$1 & 128 & 32$\times$32$\times$128 \\
\multicolumn{6}{l}{2$\times$ Bilinear Upsample} \\
64$\times$64$\times$128 & Conv & 3$\times$3 & 1$\times$1 & 64 & 64$\times$64$\times$64 \\
\multicolumn{6}{l}{2$\times$ Bilinear Upsample} \\
128$\times$128$\times$64 & Conv & 5$\times$5 & 1$\times$1 & 3 & 128$\times$128$\times$3\\\hline
\end{tabular}
\caption{Architecture of the texture decoder. Each fully connected (FC) and convolution layer is followed by a ReLU  \jp{non-linearity}, while only the last convolution layer is followed by a tanh non-linearity. The length of the input latent vector depends on the mode of the experiment.}
\label{tab_decoder}
\end{table*}
%=============== Architecture of Decoder=============

%==========Architecture of Lighting Encoder======
\begin{table*}[]
\centering
\begin{tabular}{llllll}\hline\hline
Input & Type & Kernel & Stride & Channels & Outputs \\\hline
\multicolumn{6}{l}{\begin{tabular}[c]{@{}l@{}}Input (RGB):\\ =128$\times$128$\times$3\end{tabular}} \\\hline
Input & Conv & 5$\times$5 & 2$\times$2 & 128 & 64$\times$64$\times$128 \\
64$\times$64$\times$128 & Conv & 5$\times$5 & 2$\times$2 & 256 & 32$\times$32$\times$256 \\
32$\times$32$\times$256 & Conv & 5$\times$5 & 2$\times$2 & 512 & 16$\times$16$\times$512 \\
16$\times$16$\times$512 & Conv & 5$\times$5 & 2$\times$2 & 1024 & 8$\times$8$\times$1024 \\
8$\times$8$\times$1024 & Conv & 5$\times$5 & 2$\times$2 & 2048 & 4$\times$4$\times$2048 \\
4$\times$4$\times$2048 & FC & - & - & - & 2\\ \hline
\end{tabular}
\caption{Architecture of the encoder for the lighting atlas (AR mode uses a similar encoder for previously predicted atlas). The encoder input is an RGB image and output is a latent vector. Each convolution layer is followed by a ReLU \jp{non-linearity}. The last fully-connected (FC) layer is followed by a tanh non-linearity.}
\label{tab_encoder_lighting}
\end{table*}
%============= Architecture of lighting encoder==========
{\small
\bibliographystyle{ieee_fullname}
\bibliography{egbib}
}